\documentclass{article}

\usepackage{microtype}
\usepackage{graphicx}
\usepackage{subfigure}
\usepackage{booktabs} 
\usepackage{wrapfig,lipsum,booktabs}
\usepackage{caption}
\usepackage{float}
\usepackage{placeins}
\usepackage{svg}
 \usepackage{enumitem}

\usepackage{hyperref}

\usepackage[accepted]{icml2023}

\usepackage{amsmath}
\usepackage{amssymb}
\usepackage{mathtools}
\usepackage{amsthm}
\usepackage[noabbrev,capitalize]{cleveref}

\usepackage[capitalize,noabbrev]{cleveref}

\theoremstyle{plain}

\theoremstyle{definition}

\theoremstyle{remark}

\newcommand{\ignore}[1]{}

\usepackage[textsize=tiny]{todonotes}

\begin{document}

\twocolumn[
\icmltitle{TabDDPM: Modelling Tabular Data with Diffusion Models}




\begin{icmlauthorlist}
\icmlauthor{Akim Kotelnikov}{HSE,Yandex}
\icmlauthor{Dmitry Baranchuk}{Yandex}
\icmlauthor{Ivan Rubachev}{HSE,Yandex}
\icmlauthor{Artem Babenko}{Yandex}
\end{icmlauthorlist}

\icmlaffiliation{HSE}{HSE university, Moscow, Russia}
\icmlaffiliation{Yandex}{Yandex, Moscow, Russia}

\icmlcorrespondingauthor{Akim Kotelnikov}{ya@akotelnikov.ru}

\icmlkeywords{tabular data, diffusion models, Machine Learning, ICML}

\vskip 0.3in
]



\printAffiliationsAndNotice{}  

\begin{abstract}
Denoising diffusion probabilistic models are becoming the leading generative modeling paradigm for many important data modalities. Being the most prevalent in the computer vision community, diffusion models have recently gained some attention in other domains, including speech, NLP, and graph-like data. In this work, we investigate if the framework of diffusion models can be advantageous for general tabular problems, where data points are typically represented by vectors of heterogeneous features. The inherent heterogeneity of tabular data makes it quite challenging for accurate modeling since the individual features can be of a completely different nature, i.e., some of them can be continuous and some can be discrete. To address such data types, we introduce TabDDPM --- a diffusion model that can be universally applied to any tabular dataset and handles any feature types. We extensively evaluate TabDDPM on a wide set of benchmarks and demonstrate its superiority over existing GAN/VAE alternatives, which is consistent with the advantage of diffusion models in other fields. 
The source code of TabDDPM is available at \href{https://github.com/yandex-research/tab-ddpm}{\textcolor{magenta}{GitHub}}.

\end{abstract}

\section{Introduction}
\label{sect:intro}

Denoising diffusion probabilistic models (DDPM) ~\citep{sohl2015deep, ho2020denoising} have recently become an object of great research interest in the generative modeling community since they often outperform the alternative approaches both in terms of the realism of individual samples and their diversity ~\citep{dhariwal2021diffusion}. The most impressive successes of DDPM were demonstrated in the domain of natural images \citep{dhariwal2021diffusion, saharia2022photorealistic, rombach2022high}, where the advantages of diffusion models are successfully exploited in applications, such as colorization ~\citep{song2020score}, inpainting ~\citep{song2020score}, segmentation \cite{baranchuk2021label}, super-resolution ~\citep{saharia2021image, li2021srdiff}, semantic editing ~\citep{meng2021sdedit} and others. Beyond computer vision, the DDPM framework is also investigated in other fields, such as NLP \citep{austin2021structured, li2022diffusion}, waveform signal processing \citep{kong2020diffwave, chen2020wavegrad}, molecular graphs \citep{jing2022torsional, hoogeboom2022equivariant}, time series \citep{tashiro2021csdi}, testifying the universality of diffusion models across a wide range of problems.

Our work aims to investigate if the universality of DDPM can be extended to the case of general tabular problems, which are ubiquitous in various industrial applications that include data described by a set of heterogeneous features. For many such applications, the demand for high-quality generative models is especially acute because of the modern privacy regulations, like GDPR, which prevent publishing real user data, while the synthetic data produced by generative models can be shared. 
However, training a high-quality model of tabular data can be more challenging than in computer vision or NLP due to the heterogeneity of individual features and relatively small sizes of typical tabular datasets. This paper shows that despite these two intricacies, the diffusion models can successfully approximate typical distributions of tabular data, leading to state-of-the-art performance on most of the benchmarks.
In more detail, the main contributions of this work are the following:
\vspace{-2mm}

\begin{enumerate}[leftmargin=15pt]
    \item We introduce TabDDPM --- a simple design of DDPM for tabular problems that can be applied to any tabular task and work with mixed data types, including numerical and categorical features. 
    
    \item We demonstrate that TabDDPM outperforms the alternative approaches designed for tabular data, including GAN-based and VAE-based methods, and illustrate the sources of this advantage on several datasets.

    \item We observe that shallow interpolation-based methods, e.g., SMOTE~\cite{chawla2002smote}, produce surprisingly effective synthetic data that provides competitively high ML efficiency.
    Compared with SMOTE, we show that TabDDPM's data is preferable for privacy-concerned scenarios when synthetic data is used to substitute the real user data that cannot be shared.
    
\end{enumerate}

\section{Related Work}
\label{sect:related}

\textbf{Diffusion models} ~\citep{sohl2015deep, ho2020denoising} is a paradigm of generative modeling that aims to approximate the target distribution by the endpoint of the Markov chain, which starts from a given parametric distribution, typically a standard Gaussian. Each Markov step is performed by a deep neural network that effectively learns to invert the diffusion process with a known Gaussian kernel. \citeauthor{ho2020denoising} demonstrated the equivalence of diffusion models and score matching ~\citep{song2019generative, song2020improved}, showing them to be two different perspectives on the gradual conversion of a simple known distribution into a target distribution via the iterative denoising process.
Several recent works ~\citep{nichol2021improved, dhariwal2021diffusion} have developed more powerful model architectures as well as different advanced learning protocols, which led to the ``victory'' of DDPM over GANs in terms of generative quality and diversity in the computer vision field. In this work, we demonstrate that one can also successfully use diffusion models for tabular problems.

\textbf{Generative models for tabular problems} are currently an active research direction in the machine learning community 
since high-quality synthetic data is in great demand for many tabular tasks. First, the tabular datasets are often limited in size, unlike in vision or NLP problems, for which massive ``extra'' data is available on the Internet. Second, proper synthetic datasets do not contain actual user data. Therefore, they are not subject to GDPR-like regulations and can be publicly shared without violating anonymity. 
The recent works have developed a large number of models, including tabular VAEs \citep{xu2019modeling} and GAN-based approaches \citep{xu2019modeling, engelmann2021conditional, jordon2018pate, fan2020relational, torfi2022differentially, zhao2021ctab, kim2021oct, zhang2021ganblr, nock2022generative, wen2022causal}. By extensive evaluations on a large number of public benchmarks, we show that TabDDPM surpasses the existing alternatives, often by a large margin.

\begin{figure*}[t]
    \centering
    \caption{TabDDPM scheme for classification problems; $t$, $y$ and $\ell$ denote a diffusion timestep, a class label, and logits, respectively.}
    \includegraphics[width=0.75\linewidth]{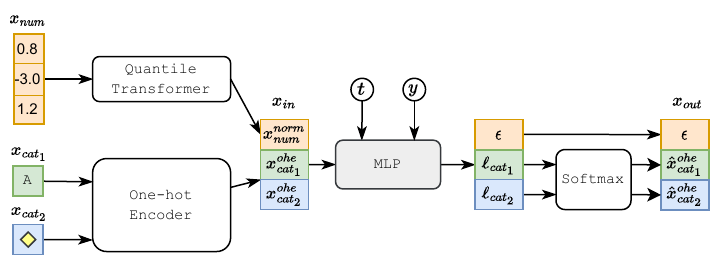}
    \label{fig:pipeline}
\end{figure*}

\textbf{``Shallow'' synthetics generation.} Unlike unstructured images or natural texts, tabular data is typically structured, i.e., the individual features are often interpretable and it is unclear if their modeling requires several layers of ``deep'' architectures. Therefore, the simple interpolation techniques, like SMOTE \citep{chawla2002smote} (originally proposed to address class imbalance) can serve as simple and powerful solutions as demonstrated in \citep{camino2020oversampling}, where SMOTE is shown to outperform tabular GANs for minor class oversampling. In the experiments, we demonstrate the advantage of TabDDPM's synthetics over synthetics produced with interpolation techniques from the privacy-preserving perspective.

\section{Background}


\textbf{Diffusion models}~\citep{sohl2015deep, ho2020denoising} are likelihood-based generative models that handle the data through forward and reverse Markov processes. The forward process $q\left(x_{1:T} | x_{0}\right){=}\prod_{t=1}^Tq\left(x_t | x_{t - 1}\right)$ gradually adds noise to an initial sample $x_0$ from the data distribution $q\left(x_0\right)$ sampling noise from the predefined distributions $q\left(x_t | x_{t - 1}\right)$ with variances $\left\{\beta_1, ..., \beta_T\right\}$. 

The reverse process $p\left(x_{0:T}\right){=}\prod_{t=1}^Tp\left(x_{t-1} | x_{t}\right)$ gradually denoises a latent variable $x_T{\sim}q\left(x_T\right)$ and allows generating new data samples from $q(x_0)$. Distributions $p\left(x_{t - 1} | x_t\right)$ are usually unknown and approximated by a neural network with parameters $\theta$. These parameters are learned from the data by optimizing a variational lower bound: 
\begingroup
\small
\begin{equation}
\label{eq:elbo}
\begin{aligned}
    & \log q\left(x_0\right) \geq \mathbb{E}_{q\left(x_0\right)} \big[\underbrace{\log p_{\theta}\left(x_0 | x_1\right)}_{L_0} - \underbrace{KL\left(q\left(x_T|x_0\right) | q\left(x_T\right)\right)}_{L_T} - \\ 
    & -\sum_{t = 2}^T \underbrace{KL\left(q\left(x_{t - 1}|x_t, x_0\right) | p_{\theta}\left(x_{t - 1} | x_t\right)\right)}_{L_t}\big]
    \end{aligned}
\end{equation}
\endgroup

\textbf{Gaussian diffusion models} operate in continuous spaces $\left(x_t \in \mathbb{R}^n\right)$ where forward and reverse processes are characterized by Gaussian distributions:
\begin{align*}
    & q\left(x_t | x_{t - 1}\right) := \mathcal{N}\left(x_t; \sqrt{1 - \beta_t}x_{t - 1}, \beta_t I\right) \\
    & q\left(x_T\right) := \mathcal{N}\left(x_T; 0, I\right) \\
    & p_{\theta}\left(x_{t - 1}| x_t\right):= \mathcal{N}\left(x_{t - 1}; \mu_{\theta}\left(x_t, t\right), \Sigma_{\theta}\left(x_t, t\right)\right)
\end{align*}

~\citet{ho2020denoising} suggest using diagonal $\Sigma_{\theta}\left(x_t, t\right)$ with a constant $\sigma_t$ and computing $\mu_{\theta}\left(x_t, t\right)$ as a function of $x_t$ and $\epsilon_{\theta}(x_t, t)$:
\[ 
    \mu_{\theta}\left(x_t, t\right) = \frac{1}{\sqrt{\alpha_t}}\left(x_t -\frac{\beta_t}{\sqrt{1 - \bar{\alpha}_t}}\epsilon_{\theta}\left(x_t, t\right)\right)
\]
where $\alpha_t := 1- \beta_t, \ \bar{\alpha}_t := \prod_{i \leq t} \alpha_i$ and $\epsilon_{\theta}(x_t, t)$ predicts a ``groundtruth'' noise component $\epsilon$ for the noisy data sample $x_t$. In practice, the objective \autoref{eq:elbo} can be simplified to the sum of mean-squared errors between $\epsilon_{\theta}(x_t, t)$ and $\epsilon$ over all timesteps $t$:

\begin{equation}
    \label{eq:mse_simple}
    L^{\text{simple}}_t = \mathbb{E}_{x_0, \epsilon, t}\|\epsilon - \epsilon_{\theta}(x_t, t)\|^2_2
\end{equation}




\textbf{Multinomial diffusion models} \citep{hoogeboom2021argmax} are designed to generate categorical data where $x_{t} \in \{0, 1\}^K$ is a one-hot encoded categorical variable with $K$ values. The multinomial forward diffusion process defines $q\left(x_t | x_{t - 1}\right)$ as a categorical distribution that corrupts the data by uniform noise over $K$ classes:
\begin{align*}
    & q(x_t | x_{t - 1}) := \text{Cat}\left(x_t; \left(1 - \beta_t\right)x_{t - 1} + \beta_t / K\right) \\
    & q\left(x_T\right) := \text{Cat}\left(x_T; 1/K\right) \\
    & q\left(x_t | x_{0}\right) = \text{Cat}\left(x_t; \bar{\alpha}_t x_{0} + \left(1 - \bar{\alpha}_t\right)/ K\right)
\end{align*}
From the equations above, the posterior $q(x_{t - 1}| x_t, x_0)$ can be derived:
\begin{equation*}
    \label{eq:multinom_loss}
    \begin{aligned}
        q\left(x_{t - 1} | x_t, x_0\right) = \text{Cat}\left(x_{t - 1};  \pi / \sum^{K}_{k = 1} \pi_k\right), \text{ where } \\[0.75ex] \pi = [\alpha_t x_t + (1 - \alpha_t) / K] \odot [\bar{\alpha}_{t - 1}x_0 + (1 - \bar{\alpha}_{t - 1})/ K] 
    \end{aligned}
\end{equation*}

The reverse distribution $p_{\theta}\left(x_{t - 1} | x_t\right)$ is parameterized as $q\left(x_{t - 1} | x_t, \hat{x}_0(x_t, t)\right)$, where $\hat{x}_0$ is predicted by a neural network. Then, the model is trained to maximize the variational lower bound \autoref{eq:elbo}.


\section{TabDDPM}
\label{sect:method}
In this section, we describe the design of TabDDPM as well as its main hyperparameters, which affect the model's effectiveness.

\textbf{TabDDPM} uses the multinomial diffusion to model the categorical and binary features, and the Gaussian diffusion to model the numerical ones. In more detail, for a tabular data sample $x = \left[x_{\text{num}}, x_{\text{cat}_1}, ..., x_{\text{cat}_C}\right]$, that consists of $N_{\text{num}}$ numerical features $x_{\text{num}} \in \mathbb{R}^{N_{\text{num}}}$ and $C$ categorical features $x_{\text{cat}_i}$ with $K_{i}$ categories each, our model takes one-hot encoded versions of categorical features as an input (i.e. $x^{\text{ohe}}_{\text{cat}_i} \in \{0, 1\}^{K_i}$) and normalized numerical features. Therefore, the input $x_{0}$ has a dimensionality of $\left(N_{\text{num}} + \sum K_i\right)$. For preprocessing, we use the gaussian quantile transformation from the scikit-learn library \citep{scikit-learn}. Each categorical feature is handled by a separate forward diffusion process, i.e., the noise components for all features are sampled independently. 
The reverse diffusion step in TabDDPM is modeled by a multi-layer neural network that
has an output of the same dimensionality as $x_0$, where the first $N_{\text{num}}$ coordinates are the predictions of $\epsilon$ for the Gaussian diffusion and the rest are the predictions of $x^{\text{ohe}}_{\text{cat}_i}$ for the multinomial diffusions.

The TabDDPM model for the classification problems is schematically presented on \autoref{fig:pipeline}. The model is trained by minimizing a sum of mean-squared error $L^{\text{simple}}_t$ (\autoref{eq:mse_simple}) for the Gaussian diffusion term and the KL divergences $L^{i}_t$ for each multinomial diffusion term (\autoref{eq:elbo}). The total loss of multinomial diffusions is additionally divided by the number of categorical features.
\begin{equation}
    L^{\text{TabDDPM}}_t = L^{\text{simple}}_t + \frac{\sum_{i \leq C} L^{i}_t}{C}
\end{equation}
For classification datasets, we use a class-conditional model, i.e., $p_{\theta}(x_{t - 1} | x_t, y)$ is learned. For regression datasets, we consider a target value as an additional numerical feature, and the joint distribution is learned.

To model the reverse process, we use a simple MLP architecture adapted from \citep{gorishniy2021revisiting}:
\begin{equation} \label{eq:mlp}
\begin{aligned}
    & \texttt{MLP}(x) = \texttt{Linear} \left(
        \texttt{MLPBlock} \left( \ldots \left( \texttt{MLPBlock}(x) \right) \right)
    \right) \\
    & \texttt{MLPBlock}(x) = \texttt{Dropout}(\texttt{ReLU}(\texttt{Linear}(x)))
\end{aligned}
\end{equation}

As in \citep{nichol2021improved, dhariwal2021diffusion}, a tabular input $x_{in}$, a timestep $t$, and a class label $y$ are processed as follows:
\begin{equation} \label{eq:process_inputs}
\begin{aligned}
    & t\_emb = \texttt{Linear}(\texttt{SiLU}(\texttt{Linear}(\texttt{SinTimeEmb}(t)))) \\
    & y\_emb = \texttt{Embedding}(y) \\
    & x = \texttt{Linear}(x_{in}) + t\_emb + y\_emb
\end{aligned}
\end{equation}

where \texttt{SinTimeEmb} refers to a sinusoidal time embedding as in \citep{nichol2021improved, dhariwal2021diffusion} with a dimension of 128. All \texttt{Linear} layers in \autoref{eq:process_inputs} have a fixed projection dimension 128.

\begin{table}[t] 
\centering
\caption{The list of main hyperparameters for TabDDPM.}
\vspace{1mm}
\small
\begin{tabular}{ll}
    \toprule
    Hyperparameter & Search space \\
    \midrule
    Learning rate & $\mathrm{LogUniform}[0.00001, 0.003]$ \\
    Batch size &  $\mathrm{Cat}\{256, 4096\}$ \\
    Diffusion timesteps & $\mathrm{Cat}\{100, 1000\}$ \\
    Training iterations &  $\mathrm{Cat}\{5000, 10000, 20000\}$ \\
    \# MLP layers &  $\mathrm{Int}\{2, 4, 6, 8\}$ \\
    MLP width of layers&  $\mathrm{Int}\{128, 256, 512, 1024\}$ \\
    Proportion of samples &  $\mathrm{Float}\{0.25, 0.5, 1, 2, 4, 8\}$ \\
    \midrule
    Dropout & 0.0 \\
    Scheduler & cosine \citep{nichol2021improved} \\
    Gaussian diffusion loss & MSE \\
    \midrule
    Number of tuning trials & 50 \\
    \bottomrule
\end{tabular}
\label{tab:tabddpm-space}
\end{table}

\textbf{Hyperparameters} in TabDDPM are essential since, in the experiments, we observed them having a strong influence on the model effectiveness. \autoref{tab:tabddpm-space} lists the main hyperparameters and the search spaces for each of them, which we recommend using. The process of tuning is described in detail in the experimental section.

\begin{table}[t]
    \centering
    \caption{Details on the datasets used in the evaluation.}
    \vspace{-2mm}
    \setlength\tabcolsep{2pt}
    \resizebox{0.49\textwidth}{!}{\begin{tabular}{lcccccccc}
\toprule
Abbr & Name & \# Train & \# Validation & \# Test & \# Num & \# Cat & Task type \\
\midrule
AB & Abalone & $2672$ & $669$ & $836$ & $7$ & $1$ & Regression \\
AD & Adult & $26048$ & $6513$ & $16281$ & $6$ & $8$ & Binclass \\
BU & Buddy & $12053$ & $3014$ & $3767$ & $4$ & $5$ & Multiclass \\
CA & California Housing & $13209$ & $3303$ & $4128$ & $8$ & $0$ & Regression \\
CAR & Cardio & $44800$ & $11200$ & $14000$ & $5$ & $6$ & Binclass \\
CH & Churn Modeling & $6400$ & $1600$ & $2000$ & $7$ & $4$ & Binclass \\
DE & Default & $19200$ & $4800$ & $6000$ & $20$ & $3$ & Binclass \\
DI & Diabetes & $491$ & $123$ & $154$ & $8$ & $0$ & Binclass \\
FB & Facebook Comm. Vol. & $157638$ & $19722$ & $19720$ & $50$ & $1$ & Regression  \\
GE & Gesture Phase & $6318$ & $1580$ & $1975$ & $32$ & $0$ & Multiclass  \\
HI & Higgs Small & $62751$ & $15688$ & $19610$ & $28$ & $0$ & Binclass  \\
HO & House 16H & $14581$ & $3646$ & $4557$ & $16$ & $0$ & Regression  \\
IN & Insurance & $856$ & $214$ & $268$ & $3$ & $3$ & Regression  \\
KI & King & $13832$ & $3458$ & $4323$ & $17$ & $3$ & Regression  \\
MI & MiniBooNE & $83240$ & $20811$ & $26013$ & $50$ & $0$ & Binclass \\
WI & Wilt & $3096$ & $775$ & $968$ & $5$ & $0$ & Binclass \\

\bottomrule
\end{tabular}}
    \label{tab:dataset_info}
\end{table}

\section{Experiments}
\label{sect:experiments}
In this section, we extensively evaluate TabDDPM against existing alternatives.

\textbf{Datasets.} For systematic investigation of the performance of tabular generative models, we consider a diverse set of $15$ real-world public datasets. These datasets have various sizes, nature, number of features, and their distributions. Most datasets were previously used for tabular model evaluation in \citep{zhao2021ctab, gorishniy2021revisiting}. The full list of datasets and their properties are presented in \autoref{tab:dataset_info}.

\textbf{Baselines.} Since the number of generative models proposed for tabular data is enormous, we evaluate TabDDPM only against the leading approaches from each paradigm of generative modeling. Also, we consider only the baselines with the published source code.

\begin{itemize}
    \item \textbf{TVAE}~\citep{xu2019modeling} --- the state-of-the-art variational auto-encoder for tabular data generation. To the best of our knowledge, there are no alternative VAE-like models that outperform TVAE and have public source code.
    \item \textbf{CTGAN}~\citep{xu2019modeling} --- arguably the most popular and well-known GAN-based model for synthetic data generation.
    \item \textbf{CTABGAN}~\citep{zhao2021ctab} --- a recent GAN-based model that is shown to outperform the existing tabular GANs on a diverse set of benchmarks. This approach cannot handle regression tasks.
    \item \textbf{CTABGAN+}~\citep{zhao2022ctab} --- an extension of the \textbf{CTABGAN} model that was published in the very recent preprint. We are unaware of the GAN-based model for tabular data proposed after \textbf{CTABGAN+} and has a public source code.
    \item \textbf{SMOTE}~\citep{chawla2002smote} --- a ``shallow'' interpolation-based method that "generates" a synthetic point as a convex combination of a real data point and its $k$-th nearest neighbor from the dataset. This method was originally proposed for minor class oversampling. Here, we generalize it to synthetic data generation as a simple sanity check, i.e., a new synthetic sample is "generated" by interpolating two samples from the same class. For regression problems, we split data into two classes by the median of the target variable.
\end{itemize}

\textbf{Evaluation measure.} Our primary evaluation measure is \textit{machine learning (ML) efficiency} (or utility) \citep{xu2019modeling}. In more detail, ML efficiency quantifies the performance of classification or regression models trained on synthetic data and evaluated on the real test set. Intuitively, models trained on high-quality synthetics should be competitive (or even superior) to models trained on real data. In our experiments, we use two evaluation protocols to compute ML efficiency. In the first protocol, which is more common in the literature \citep{xu2019modeling, zhao2021ctab, kim2022sos}, we compute an average efficiency w.r.t. a set of diverse ML models (logistic regression, decision tree, and others). In the second protocol, we evaluate ML efficiency only w.r.t. the CatBoost model \citep{prokhorenkova2018catboost}, which is arguably the leading GBDT implementation providing state-of-the-art performance on tabular tasks \cite{gorishniy2021revisiting}. In our experiments in \autoref{subsec:ml-eff}, we show that it is crucial to use the second protocol, while the first one can often be misleading.

\begin{figure*}[th!]
\centering
\caption{The individual feature distributions for the real data and the data generated by TabDDPM, CTABGAN+, and TVAE. TabDDPM produces more realistic feature distributions than alternatives in most cases.}
\includegraphics[width=0.80\linewidth]{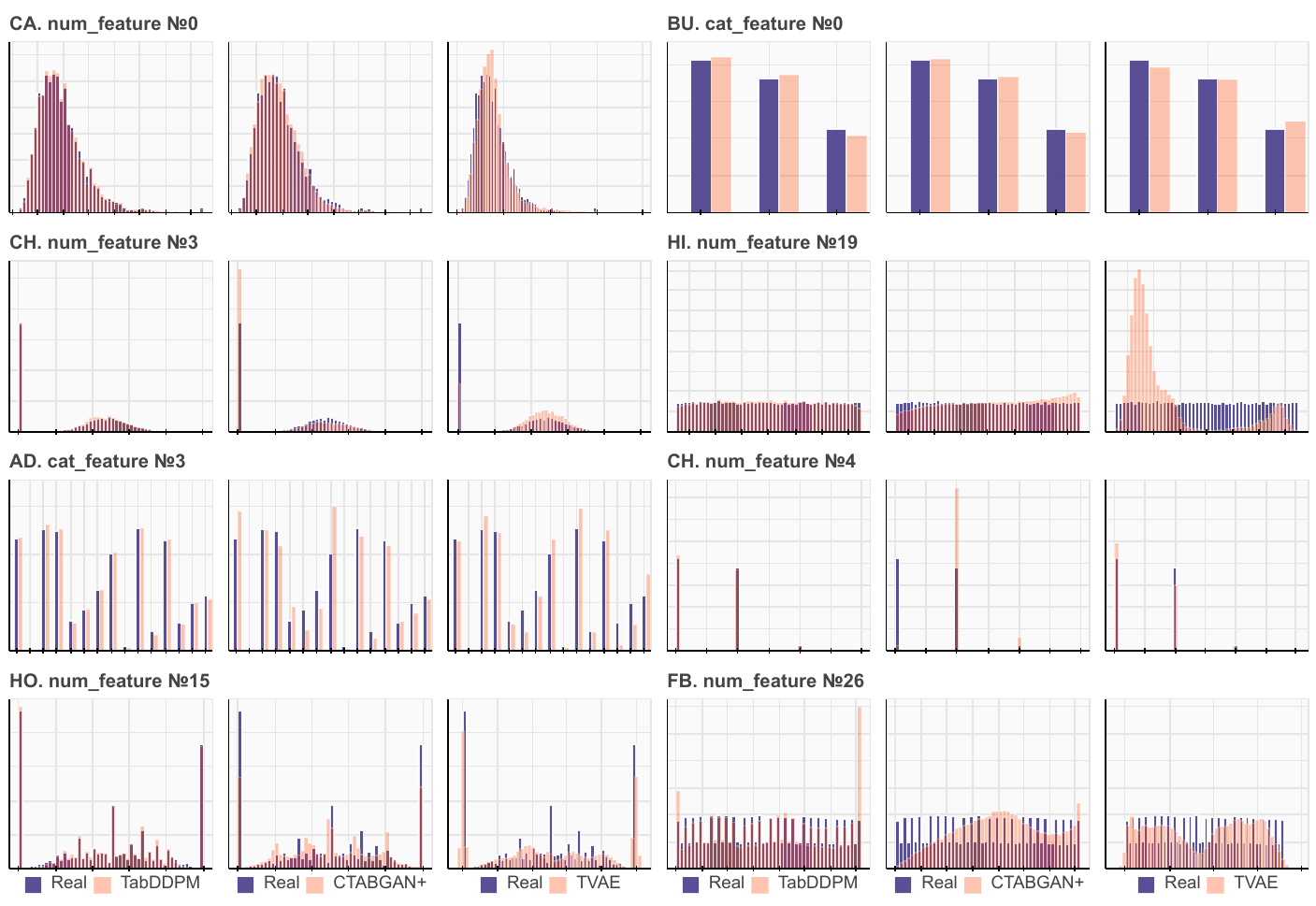}
\label{fig:features}
\end{figure*}

\begin{figure}[th!]
\centering
\caption{Absolute difference between correlation matrices computed on real and synthetic datasets. A more intensive red color indicates a higher difference between the real and synthetic correlation values. In most cases, TabDDPM captures feature correlations better.}
\includegraphics[width=0.99\linewidth]{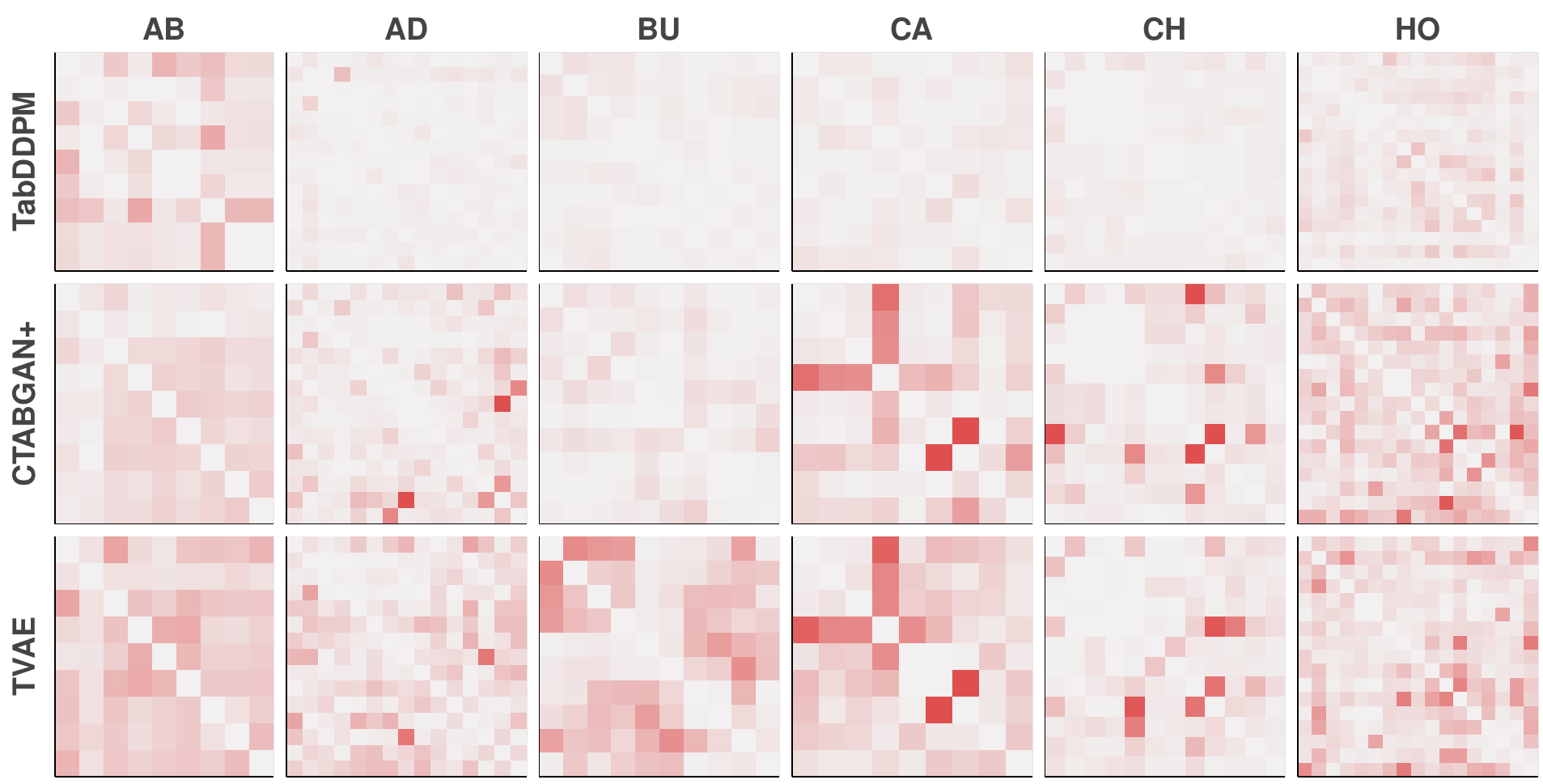}
\label{fig:correlations}
\end{figure}

\textbf{Tuning process.} To tune the hyperparameters of TabDDPM and the baselines, we use the Optuna library \citep{akiba2019optuna}. The tuning process is guided by the values of the ML efficiency (w.r.t. Catboost) of the generated synthetic data on a hold-out validation dataset (the score is averaged over five different sampling seeds). The search spaces for all hyperparameters of TabDDPM are reported in \autoref{tab:tabddpm-space} (for baselines --- in \autoref{sec:spaces}). Additionally, we demonstrate that tuning the hyperparameters using the CatBoost guidance does not introduce any sort of ``Catboost-biasedness'', and the Catboost-tuned TabDDPM produces synthetics that are also superior for other models, like MLP. These results are reported in \autoref{app:mlp-eval-tune}.

\subsection{Qualitative comparison}

Here, we qualitatively investigate the ability of TabDDPM to model the individual and joint feature distributions compared with the TVAE and CTABGAN+ baselines. In particular, for each dataset, we produce synthetic datasets from TabDDPM, TVAE, and CTABGAN+ of the same size as a real train set in a particular dataset. For classification datasets, each class is sampled according to its proportion in the real dataset. Then, we visualize the typical individual feature distributions for real and synthetic data in \autoref{fig:features}. For completeness, the features of different types and distributions are presented. 

In most cases, TabDDPM produces more realistic feature distributions compared with TVAE and CTABGAN+. The advantage is more pronounced (1) for numerical features, which are uniformly distributed, (2) for categorical features with high cardinality, and  (3) for mixed-type features that combine continuous and discrete distributions. Then, we also visualize the differences between the correlation matrices computed on real and synthetic data for different datasets, see \autoref{fig:correlations}. To compute the correlation matrices, we use the Pearson correlation coefficient for numerical-numerical correlations, the correlation ratio for categorical-numerical cases, and  Theil's U statistic between categorical features. In comparison with CTABGAN+ and TVAE, TabDDPM generates synthetic datasets with more realistic pairwise correlations. These illustrations indicate that our TabDDPM model is more flexible than alternatives and produces superior synthetic data. We also follow \cite{zhao2021ctab} and measure the Wasserstein distance between numerical features and the Jensen–Shannon divergence between categorical ones. Additionally, we report an L2 distance between correlation matrices (quantitative results for \autoref{fig:correlations}). The results are presented in \autoref{tab:ranks} as an average rank across all datasets (lower is better). Lower rank indicates lower WD, JS divergence and L2 distance. The exact numbers can be found in \autoref{app:additional}.

\begin{table}
    \centering
    \caption{Average ranks (lower is better) over all datasets in terms of Wasserstein distance (WD) between numerical features, Jensen–Shannon divergence between categorical features and L2 distance between correlation matrices. Distances are calculated between generated data and real data.}
    \vspace*{1mm}
    \resizebox{0.4\textwidth}{!}{
    \begin{tabular}{l|c|c|c}
    \toprule
            {} &  WD (Num.) & JS (Cat.) & L2 (Corr. matrix) \\
    \midrule
         CTGAN & 3.33 & 4.77 & 3.47 \\
         TVAE &  4.20 & 3.92 & 4.40 \\ 
         CTABGAN+ & 3.87 & 2.54 & 3.40 \\
         SMOTE & \textbf{1.67} & 2.15 & 2.00 \\ 
         TabDDPM & 1.93 & \textbf{1.62} & \textbf{1.73} \\
         \bottomrule
    \end{tabular}}

    \label{tab:ranks}
\end{table}

\subsection{Machine Learning efficiency}
\label{subsec:ml-eff}

\begin{table*}[ht!]
    \centering
    \caption{The values of machine learning efficiency computed w.r.t. five weak classification/regression models. Negative \\ scores denote negative R2, which means that performance is worse than an optimal constant prediction.}
    \vspace{-2mm}
    \resizebox{0.88\textwidth}{!}{\begin{tabular}{lcccccccc}
\toprule
{} & AB {\tiny $\left(R2\right)$}  & AD {\tiny $\left(F1\right)$} & BU {\tiny $\left(F1\right)$} & CA {\tiny $\left(R2\right)$} & CAR {\tiny $\left(F1\right)$} & CH {\tiny $\left(F1\right)$} & DE {\tiny $\left(F1\right)$} & DI {\tiny $\left(F1\right)$} \\
\midrule
TVAE            & $0.238 \scriptscriptstyle \pm \scriptstyle .012$ & $0.742 \scriptscriptstyle \pm \scriptstyle .001$ & $0.779 \scriptscriptstyle \pm \scriptstyle .004$ & $-13.0 \scriptscriptstyle \pm \scriptstyle 1.51$ & $0.693 \scriptscriptstyle \pm \scriptstyle .002$ & $0.684 \scriptscriptstyle \pm \scriptstyle .003$ & $0.643 \scriptscriptstyle \pm \scriptstyle .003$ & $0.712 \scriptscriptstyle \pm \scriptstyle .010$ \\
CTABGAN         & -- & $0.737 \scriptscriptstyle \pm \scriptstyle .007$ & $0.786 \scriptscriptstyle \pm \scriptstyle .008$ & -- & $0.684 \scriptscriptstyle \pm \scriptstyle .003$ & $0.636 \scriptscriptstyle \pm \scriptstyle .010$ & $0.614 \scriptscriptstyle \pm \scriptstyle .007$ & $0.655 \scriptscriptstyle \pm \scriptstyle .015$ \\
CTABGAN+        & $0.316 \scriptscriptstyle \pm \scriptstyle .024$ & $0.730 \scriptscriptstyle \pm \scriptstyle .007$ & $0.837 \scriptscriptstyle \pm \scriptstyle .006$ & $-7.59 \scriptscriptstyle \pm \scriptstyle .645$ & $\mathbf{0.708 \scriptscriptstyle \pm \scriptstyle .002}$ & $0.650 \scriptscriptstyle \pm \scriptstyle .008$ & $0.648 \scriptscriptstyle \pm \scriptstyle .008$ & $\mathbf{0.727 \scriptscriptstyle \pm \scriptstyle .023}$ \\
SMOTE           & $\mathbf{0.400 \scriptscriptstyle \pm \scriptstyle .009}$ & $0.750 \scriptscriptstyle \pm \scriptstyle .004$ & $0.842 \scriptscriptstyle \pm \scriptstyle .003$ & $0.667 \scriptscriptstyle \pm \scriptstyle .006$ & $0.693 \scriptscriptstyle \pm \scriptstyle .001$ & $0.690 \scriptscriptstyle \pm \scriptstyle .003$ & $0.649 \scriptscriptstyle \pm \scriptstyle .003$ & $0.677 \scriptscriptstyle \pm \scriptstyle .013$ \\
TabDDPM         & $\mathbf{0.392 \scriptscriptstyle \pm \scriptstyle .009}$ & $\mathbf{0.758 \scriptscriptstyle \pm \scriptstyle .005}$ & $\mathbf{0.851 \scriptscriptstyle \pm \scriptstyle .003}$ & $\mathbf{0.695 \scriptscriptstyle \pm \scriptstyle .002}$ & $0.696 \scriptscriptstyle \pm \scriptstyle .001$ & $\mathbf{0.693 \scriptscriptstyle \pm \scriptstyle .003}$ & $\mathbf{0.659 \scriptscriptstyle \pm \scriptstyle .003}$ & $0.675 \scriptscriptstyle \pm \scriptstyle .011$ \\
\midrule 
Real            & $0.423 \scriptscriptstyle \pm \scriptstyle .009$ & $0.750 \scriptscriptstyle \pm \scriptstyle .006$ & $0.845 \scriptscriptstyle \pm \scriptstyle .004$ & $0.663 \scriptscriptstyle \pm \scriptstyle .002$ & $0.683 \scriptscriptstyle \pm \scriptstyle .002$ & $0.679 \scriptscriptstyle \pm \scriptstyle .003$ & $0.648 \scriptscriptstyle \pm \scriptstyle .003$ & $0.699 \scriptscriptstyle \pm \scriptstyle .012$ \\
\bottomrule
\end{tabular}}

\vspace*{0.1em}

\resizebox{0.88\textwidth}{!}{\begin{tabular}{lcccccccc}
\toprule
{} & FB {\tiny $\left(R2\right)$} & GE {\tiny $\left(F1\right)$} & HI {\tiny $\left(F1\right)$} & HO {\tiny $\left(R2\right)$} & IN {\tiny $\left(R2\right)$} & KI {\tiny $\left(R2\right)$} & MI {\tiny $\left(F1\right)$} & WI {\tiny $\left(F1\right)$} \\
\midrule
TVAE            & $\ll 0$ & $0.372 \scriptscriptstyle \pm \scriptstyle .006$ & $0.590 \scriptscriptstyle \pm \scriptstyle .004$ & $0.174 \scriptscriptstyle \pm \scriptstyle .012$ & $0.470 \scriptscriptstyle \pm \scriptstyle .024$ & $0.666 \scriptscriptstyle \pm \scriptstyle .006$ & $\mathbf{0.880 \scriptscriptstyle \pm \scriptstyle .002}$ & $0.497 \scriptscriptstyle \pm \scriptstyle .001$ \\
CTABGAN         & -- & $0.339 \scriptscriptstyle \pm \scriptstyle .009$ & $0.539 \scriptscriptstyle \pm \scriptstyle .006$ & -- & -- & -- & $0.856 \scriptscriptstyle \pm \scriptstyle .003$ & $0.656 \scriptscriptstyle \pm \scriptstyle .011$ \\
CTABGAN+        & $\ll 0$ & $0.373 \scriptscriptstyle \pm \scriptstyle .009$ & $0.598 \scriptscriptstyle \pm \scriptstyle .004$ & $0.222 \scriptscriptstyle \pm \scriptstyle .042$ & $0.669 \scriptscriptstyle \pm \scriptstyle .018$ & $0.197 \scriptscriptstyle \pm \scriptstyle .051$ & $0.867 \scriptscriptstyle \pm \scriptstyle .002$ & $0.653 \scriptscriptstyle \pm \scriptstyle .027$ \\
SMOTE           & $\mathbf{0.651 \scriptscriptstyle \pm \scriptstyle .002}$ & $\mathbf{0.478 \scriptscriptstyle \pm \scriptstyle .005}$ & $0.664 \scriptscriptstyle \pm \scriptstyle .003$ & $0.394 \scriptscriptstyle \pm \scriptstyle .006$ & $0.709 \scriptscriptstyle \pm \scriptstyle .008$ & $\mathbf{0.751 \scriptscriptstyle \pm \scriptstyle .005}$ & $0.860 \scriptscriptstyle \pm \scriptstyle .001$ & $\mathbf{0.793 \scriptscriptstyle \pm \scriptstyle .004}$ \\
TabDDPM         & $0.527 \scriptscriptstyle \pm \scriptstyle .005$ & $0.462 \scriptscriptstyle \pm \scriptstyle .005$ & $\mathbf{0.670 \scriptscriptstyle \pm \scriptstyle .002}$ & $\mathbf{0.426 \scriptscriptstyle \pm \scriptstyle .007}$ & $\mathbf{0.734 \scriptscriptstyle \pm \scriptstyle .007}$ & $0.611 \scriptscriptstyle \pm \scriptstyle .013$ & $0.850 \scriptscriptstyle \pm \scriptstyle .004$ & $\mathbf{0.792 \scriptscriptstyle \pm \scriptstyle .004}$ \\
\midrule
Real            & $0.645 \scriptscriptstyle \pm \scriptstyle .005$ & $0.431 \scriptscriptstyle \pm \scriptstyle .005$ & $0.663 \scriptscriptstyle \pm \scriptstyle .002$ & $0.415 \scriptscriptstyle \pm \scriptstyle .007$ & $0.708 \scriptscriptstyle \pm \scriptstyle .007$ & $0.768 \scriptscriptstyle \pm \scriptstyle .013$ & $0.850 \scriptscriptstyle \pm \scriptstyle .004$ & $0.684 \scriptscriptstyle \pm \scriptstyle .004$ \\
\bottomrule
\end{tabular}}
    \label{tab:ml_utility_avg}
\end{table*}
\begin{table*}[ht!]
    \centering
    \caption{The values of machine learning efficiency computed w.r.t. the state-of-the-art tuned CatBoost model.}
    \vspace{-2mm}
    \resizebox{0.88\textwidth}{!}{\begin{tabular}{lcccccccc}
\toprule
{} & AB {\tiny $\left(R2\right)$}  & AD {\tiny $\left(F1\right)$} & BU {\tiny $\left(F1\right)$} & CA {\tiny $\left(R2\right)$} & CAR {\tiny $\left(F1\right)$} & CH {\tiny $\left(F1\right)$} & DE {\tiny $\left(F1\right)$} & DI {\tiny $\left(F1\right)$} \\
\midrule
CTGAN           & $0.420 \scriptscriptstyle \pm \scriptstyle .004$ & $0.789 \scriptscriptstyle \pm \scriptstyle .001$ & $0.867 \scriptscriptstyle \pm \scriptstyle .003$ & $0.686 \scriptscriptstyle \pm \scriptstyle .003$ & $0.730 \scriptscriptstyle \pm \scriptstyle .001$ & $0.723 \scriptscriptstyle \pm \scriptstyle .006$ & $\mathbf{0.699 \scriptscriptstyle \pm \scriptstyle .002}$ & $0.459 \scriptscriptstyle \pm \scriptstyle .096$ \\

TVAE            & $0.433 \scriptscriptstyle \pm \scriptstyle .008$ & $0.781 \scriptscriptstyle \pm \scriptstyle .002$ & $0.864 \scriptscriptstyle \pm \scriptstyle .005$ & $0.752 \scriptscriptstyle \pm \scriptstyle .001$ & $0.717 \scriptscriptstyle \pm \scriptstyle .001$ & $0.732 \scriptscriptstyle \pm \scriptstyle .006$ & $0.656 \scriptscriptstyle \pm \scriptstyle .007$ & $\mathbf{0.714 \scriptscriptstyle \pm \scriptstyle .039}$ \\
CTABGAN         & -- & $0.783 \scriptscriptstyle \pm \scriptstyle .002$ & $0.855 \scriptscriptstyle \pm \scriptstyle .005$ & -- & $0.717 \scriptscriptstyle \pm \scriptstyle .001$ & $0.688 \scriptscriptstyle \pm \scriptstyle .006$ & $0.644 \scriptscriptstyle \pm \scriptstyle .011$ & $\mathbf{0.731 \scriptscriptstyle \pm \scriptstyle .022}$ \\
CTABGAN+        & $0.467 \scriptscriptstyle \pm \scriptstyle .004$ & $0.772 \scriptscriptstyle \pm \scriptstyle .003$ & $0.884 \scriptscriptstyle \pm \scriptstyle .005$ & $0.525 \scriptscriptstyle \pm \scriptstyle .004$ & $0.733 \scriptscriptstyle \pm \scriptstyle .001$ & $0.702 \scriptscriptstyle \pm \scriptstyle .012$ & $0.686 \scriptscriptstyle \pm \scriptstyle .004$ & $\mathbf{0.734 \scriptscriptstyle \pm \scriptstyle .020}$ \\
SMOTE           & $\mathbf{0.549 \scriptscriptstyle \pm \scriptstyle .005}$ & $0.791 \scriptscriptstyle \pm \scriptstyle .002$ & $0.891 \scriptscriptstyle \pm \scriptstyle .003$ & $\mathbf{0.840 \scriptscriptstyle \pm \scriptstyle .001}$ & $0.732 \scriptscriptstyle \pm \scriptstyle .001$ & $0.743 \scriptscriptstyle \pm \scriptstyle .005$ & $0.693 \scriptscriptstyle \pm \scriptstyle .003$ & $0.683 \scriptscriptstyle \pm \scriptstyle .037$ \\
TabDDPM         & $\mathbf{0.550 \scriptscriptstyle \pm \scriptstyle .010}$ & $\mathbf{0.795 \scriptscriptstyle \pm \scriptstyle .001}$ & $\mathbf{0.906 \scriptscriptstyle \pm \scriptstyle .003}$ & $0.836 \scriptscriptstyle \pm \scriptstyle .002$ & $\mathbf{0.737 \scriptscriptstyle \pm \scriptstyle .001}$ & $\mathbf{0.755 \scriptscriptstyle \pm \scriptstyle .006}$ & $0.691 \scriptscriptstyle \pm \scriptstyle .004$ & $\mathbf{0.740 \scriptscriptstyle \pm \scriptstyle .020}$ \\
\midrule
Real            & $0.556 \scriptscriptstyle \pm \scriptstyle .004$ & $0.815 \scriptscriptstyle \pm \scriptstyle .002$ & $0.906 \scriptscriptstyle \pm \scriptstyle .002$ & $0.857 \scriptscriptstyle \pm \scriptstyle .001$ & $0.738 \scriptscriptstyle \pm \scriptstyle .001$ & $0.740 \scriptscriptstyle \pm \scriptstyle .009$ & $0.688 \scriptscriptstyle \pm \scriptstyle .003$ & $0.785 \scriptscriptstyle \pm \scriptstyle .013$ \\
\bottomrule
\end{tabular}}

\vspace*{0.1em}

\resizebox{0.88\textwidth}{!}{\begin{tabular}{lcccccccc}
\toprule
{} & FB {\tiny $\left(R2\right)$} & GE {\tiny $\left(F1\right)$} & HI {\tiny $\left(F1\right)$} & HO {\tiny $\left(R2\right)$} & IN {\tiny $\left(R2\right)$} & KI {\tiny $\left(R2\right)$} & MI {\tiny $\left(F1\right)$} & WI {\tiny $\left(F1\right)$} \\
\midrule
CTGAN           & $0.443 \scriptscriptstyle \pm \scriptstyle .005$ & $0.333 \scriptscriptstyle \pm \scriptstyle .013$ & $0.575 \scriptscriptstyle \pm \scriptstyle .006$ & $0.433 \scriptscriptstyle \pm \scriptstyle .005$ & $0.745 \scriptscriptstyle \pm \scriptstyle .009$ & $0.772 \scriptscriptstyle \pm \scriptstyle .005$ & $0.783 \scriptscriptstyle \pm \scriptstyle .005$ & $0.749 \scriptscriptstyle \pm \scriptstyle .015$ \\
TVAE            & $0.685 \scriptscriptstyle \pm \scriptstyle .003$ & $0.434 \scriptscriptstyle \pm \scriptstyle .006$ & $0.638 \scriptscriptstyle \pm \scriptstyle .003$ & $0.493 \scriptscriptstyle \pm \scriptstyle .006$ & $0.784 \scriptscriptstyle \pm \scriptstyle .010$ & $0.824 \scriptscriptstyle \pm \scriptstyle .003$ & $0.912 \scriptscriptstyle \pm \scriptstyle .001$ & $0.501 \scriptscriptstyle \pm \scriptstyle .012$ \\
CTABGAN         & -- & $0.392 \scriptscriptstyle \pm \scriptstyle .006$ & $0.575 \scriptscriptstyle \pm \scriptstyle .004$ & -- & -- & -- & $0.889 \scriptscriptstyle \pm \scriptstyle .002$ & $\mathbf{0.906 \scriptscriptstyle \pm \scriptstyle .019}$ \\
CTABGAN+        & $0.509 \scriptscriptstyle \pm \scriptstyle .011$ & $0.406 \scriptscriptstyle \pm \scriptstyle .009$ & $0.664 \scriptscriptstyle \pm \scriptstyle .002$ & $0.504 \scriptscriptstyle \pm \scriptstyle .005$ & $0.797 \scriptscriptstyle \pm \scriptstyle .005$ & $0.444 \scriptscriptstyle \pm \scriptstyle .014$ & $0.892 \scriptscriptstyle \pm \scriptstyle .002$ & $0.798 \scriptscriptstyle \pm \scriptstyle .021$ \\
SMOTE           & $\mathbf{0.803 \scriptscriptstyle \pm \scriptstyle .002}$ & $\mathbf{0.658 \scriptscriptstyle \pm \scriptstyle .007}$ & $\mathbf{0.722 \scriptscriptstyle \pm \scriptstyle .001}$ & $0.662 \scriptscriptstyle \pm \scriptstyle .004$ & $\mathbf{0.812 \scriptscriptstyle \pm \scriptstyle .002}$ & $\mathbf{0.842 \scriptscriptstyle \pm \scriptstyle .004}$ & $0.932 \scriptscriptstyle \pm \scriptstyle .001$ & $\mathbf{0.913 \scriptscriptstyle \pm \scriptstyle .007}$ \\
TabDDPM         & $0.713 \scriptscriptstyle \pm \scriptstyle .002$ & $0.597 \scriptscriptstyle \pm \scriptstyle .006$ & $\mathbf{0.722 \scriptscriptstyle \pm \scriptstyle .001}$ & $\mathbf{0.677 \scriptscriptstyle \pm \scriptstyle .010}$ & $0.809 \scriptscriptstyle \pm \scriptstyle .002$ & $\mathbf{0.833 \scriptscriptstyle \pm \scriptstyle .014}$ & $\mathbf{0.936 \scriptscriptstyle \pm \scriptstyle .001}$ & $\mathbf{0.904 \scriptscriptstyle \pm \scriptstyle .009}$ \\
\midrule
Real            & $0.837 \scriptscriptstyle \pm \scriptstyle .001$ & $0.636 \scriptscriptstyle \pm \scriptstyle .007$ & $0.724 \scriptscriptstyle \pm \scriptstyle .001$ & $0.662 \scriptscriptstyle \pm \scriptstyle .003$ & $0.814 \scriptscriptstyle \pm \scriptstyle .001$ & $0.907 \scriptscriptstyle \pm \scriptstyle .002$ & $0.934 \scriptscriptstyle \pm \scriptstyle .000$ & $0.898 \scriptscriptstyle \pm \scriptstyle .006$ \\
\bottomrule
\end{tabular}}
    \label{tab:ml_utility_cb}
\end{table*}

In this section, we compare TabDDPM to alternative generative models in terms of machine learning efficiency. From each generative model, we sample a synthetic dataset with the size of a real train set in proportion from \autoref{tab:tabddpm-space}. This synthetic data is then used to train a classification/regression model, which is then evaluated using the real test set. In our experiments, classification performance is evaluated by the F1 score, and regression performance is evaluated by the R2 score. We use two protocols:
\vspace{-0.5em}
\begin{enumerate}
    \setlength\itemsep{0.1em}
    \item First, we compute average ML efficiency for a diverse set of ML models, as performed in previous works \citep{xu2019modeling, zhao2021ctab, kim2022sos}. This set includes Decision Tree, Random Forest, Logistic Regression (or Ridge Regression) and MLP models from the scikit-learn library \citep{scikit-learn} with the default hyperparameters except for: ``max-depth'' equals to $28$ for Decision Tree and Random Forest, ``maximum iterations'' equals to $500$ for Logistic and Ridge regressions, and ``maximum iterations'' equals to $100$ for MLPs.
    
    \item Second, we compute ML efficiency w.r.t. the current state-of-the-art model for tabular data. Specifically, we consider CatBoost \citep{prokhorenkova2018catboost} and MLP architecture from \citep{gorishniy2021revisiting} for evaluation. CatBoost and MLP hyperparameters are thoroughly tuned on each dataset using the search spaces from \citep{gorishniy2021revisiting}. We argue that this evaluation protocol demonstrates the practical value of synthetic data more reliably since in most real scenarios practitioners are not interested in using weak and suboptimal classifiers/regressors.
\end{enumerate}

\textbf{Main results.} The ML efficiency values computed by both protocols are presented in Tables~\ref{tab:ml_utility_avg},~\ref{tab:ml_utility_cb}. The ML efficiency for the tuned MLP is reported in \autoref{app:mlp-eval-tune}. To compute each value, we average the results over five random seeds for synthetics generation, and for each generated dataset, we average over ten random seeds for training classifiers/regressors. The key observations are described below:
\vspace{-0.5em}
\begin{itemize}
    \item In both evaluation protocols, TabDDPM significantly outperforms TVAE and CTABGAN+ on most datasets, which highlights the advantage of diffusion models for tabular data as well as demonstrated for other domains in prior works.
    
    \item The interpolation-based SMOTE method demonstrates the performance competitive to TabDDPM and often significantly outperforms the GAN/VAE approaches. Interestingly, most of the prior works on generative models for tabular data do not compare against SMOTE, while it appears to be a simple baseline, which is challenging to beat.
    
    \item While many prior works use the first evaluation protocol to compute the ML efficiency, we argue that the second one (which uses the state-of-the-art model) is more appropriate. Tables~\ref{tab:ml_utility_avg},~\ref{tab:ml_utility_cb} show that the absolute values of classification/regression performance are much lower for the first protocol, i.e., weak classifiers/regressors are substantially inferior to CatBoost on the considered benchmarks. Therefore, one can hardly use these suboptimal models instead of CatBoost and their performance values are uninformative for practitioners.
    Moreover, in the first protocol, training on synthetic data is often advantageous compared to training on real data. This creates an impression that the data produced by generative models are more valuable than the real ones. However, it is not the case when one uses the tuned ML model, as in most practical scenarios. \autoref{app:mlp-eval-tune} confirms this observation for the properly tuned MLP model.
\end{itemize}

Overall, TabDDPM provides state-of-the-art generative performance and can be used as a source of high-quality synthetic data. Interestingly, in terms of ML efficiency, a simple ``shallow'' SMOTE method is competitive to TabDDPM, which raises the question if sophisticated deep generative models are needed. In the section below, we provide an affirmative answer to this question.

\subsection{Privacy}
\label{subsec:privacy}

Here, we investigate TabDDPM in privacy-concerned settings, e.g., sharing the data without disclosure of personal or sensitive information. 
In these setups, one is interested in high-quality synthetic data that does not reveal the records from the original dataset. 

We measure the privacy of the generated data as a mean Distance to Closest Record (DCR)~\citep{zhao2021ctab}. 
Specifically, for each synthetic sample, we get the minimum L2 distance to the real records. 
Mean DCR averages these distances over all generated samples. 

Low DCR values indicate that synthetic samples essentially mimic some real datapoints and can violate privacy requirements. 
Higher DCR values denote that the generative model can produce ``new'' records rather than just near duplicates of the real data. 
Note that out-of-distribution data, e.g., random noise, will also provide high DCR.
Therefore, DCR needs to be considered along with ML efficiency together. 

\autoref{tab:privacy_full} presents the DCR values for TabDDPM, SMOTE, CTABGAN+ and TVAE. 
We observe that TabDDPM is more private than SMOTE and less private than GAN/VAE alternatives. 
We attribute this to significantly lower ML utility of GAN/VAE-based baselines.

Since SMOTE computes convex combinations of the real records, the original DCR measure can pessimize SMOTE's privacy.
To address this issue, we pretrain an MLP model on each dataset using real data. 
Then, we use this model to extract features from synthetic data and measure DCR in the latent space of the pretrained model. 
\autoref{tab:privacy_full_mlp} provides mean DCR values on MLP features. 
The results are mostly consistent with \autoref{tab:privacy_full} and do not alter our conclusions. 

\begin{figure}[!ht]
    \vspace{-0.25cm}
    \centering
    \caption{Histograms of minimal synthetic-to-real distances for TabDDPM and SMOTE. SMOTE values are concentrated around zero and, thus, SMOTE generates less private synthetic data.}
    \includegraphics[width=0.85\linewidth]{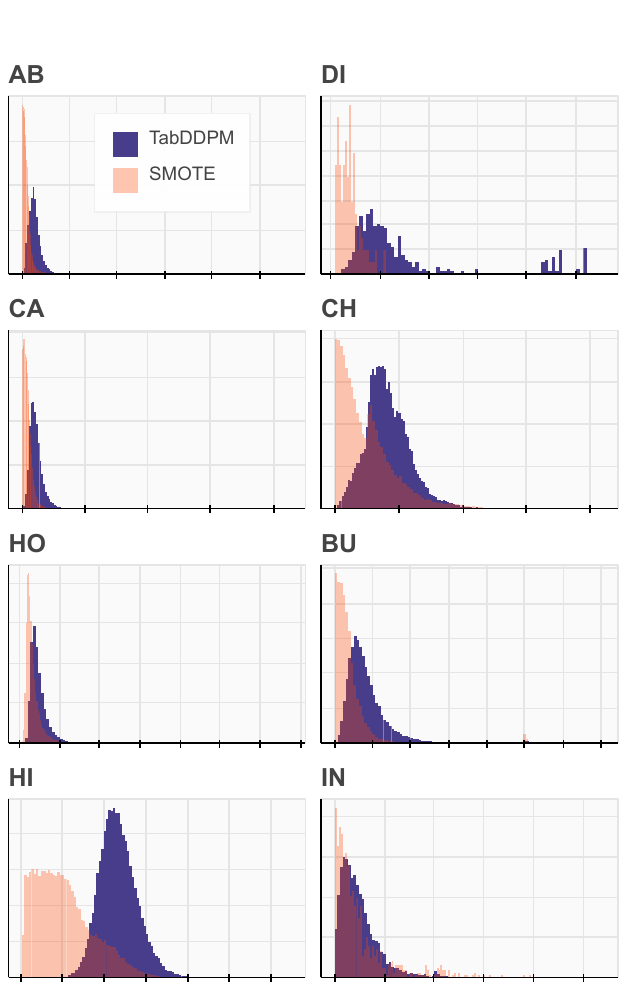}
    \label{fig:privacy_hists}
\end{figure}

We also visualize histograms of the minimal synthetic-to-real distances in \autoref{fig:privacy_hists}. For SMOTE, most distance values are concentrated around zero, while TabDDPM samples are noticeably farther from real datapoints. 

In addition, following \cite{chen2020gan, lee2021invertible}, we measure a success rate of a full black-box privacy attack (see \autoref{tab:bb_vert}). 
The attack aims to infer whether a record belongs to its original training data. 
The results show that TabDDPM is more resistant to this full black-box attack than SMOTE. All these experiments confirm that TabDDPM significantly outperforms SMOTE in privacy-concerned scenarios and still provides state-of-the-art ML efficiency. 


\begin{table}[!ht]
    \centering
    \caption{Success rate of a full black-box privacy attack in terms of ROCAUC. A higher score indicates the higher success of attack. TabDDPM is significantly more robust than SMOTE.}
    {\setlength{\tabcolsep}{4pt}
        {\footnotesize \begin{tabular}{lcc}
\toprule
{} & SMOTE & TabDDPM \\
\midrule
AB & $0.967$ & ${0.505}$ \\
AD & $0.619$ & ${0.511}$ \\
BU & $0.710$ & ${0.569}$ \\
CA & $0.986$ & ${0.516}$ \\
CAR & $0.721$ & ${0.506}$ \\
CH & $0.891$ & ${0.721}$ \\
DE & $0.679$ & ${0.497}$ \\
DI & $0.610$ & ${0.510}$ \\
GE & $0.864$ & ${0.533}$ \\
HI & $0.999$ & ${0.527}$ \\
HO & $0.826$ & ${0.546}$ \\
IN & $0.712$ & ${0.868}$ \\
KI & $0.748$ & ${0.517}$ \\
MI & $0.990$ & ${0.500}$ \\
WI & $0.954$ & $0.516$ \\
\bottomrule
\end{tabular}
}
    }
    \label{tab:bb_vert}
\end{table}

\begin{table}[!ht]
    \vspace{-0.5cm}
    \centering
    \caption{Comparison in terms of mean Distance to Closest Record (DCR) (higher is better). TabDDPM provides better DCR values compared with SMOTE but underperforms compared with TVAE and CTABGAN+. We attribute this to significantly lower ML efficiency of GAN/VAE-based alternatives.}
    \setlength\tabcolsep{3pt}
    \resizebox{0.48\textwidth}{!}{\begin{tabular}{lcccccccc}
\toprule
{} & AB & AD & BU & CA & CAR & CH & DE & DI \\
\midrule
TVAE            & $0.088$ & $0.220$ & $0.226$ & $0.056$ & $0.010$ & $0.241$ & $0.096$ & $0.146$ \\
CTABGAN+        & $0.081$ & $0.400$ & $0.242$ & $0.070$ & $0.020$ & $0.235$ & $0.131$ & $0.204$ \\
SMOTE           & $0.018$ & $0.082$ & $0.080$ & $0.016$ & $0.007$ & $0.099$ & $0.054$ & $0.074$ \\
TabDDPM         & $0.061$ & $0.295$ & $0.168$ & $0.045$ & $0.016$ & $0.166$ & $0.061$ & $0.308$ \\
\bottomrule
\end{tabular}}

\resizebox{0.48\textwidth}{!}{\begin{tabular}{lcccccccc}
\toprule
{} & FB & GE & HI & HO & IN & KI & MI & WI \\
\midrule
TVAE            & $1.418$ & $0.171$ & $0.497$ & $0.127$ & $0.102$ & $0.200$ & $0.025$ & $0.020$ \\
CTABGAN+        & $0.666$ & $0.169$ & $0.533$ & $0.129$ & $0.124$ & $0.390$ & $10.761$ & $0.027$ \\
SMOTE           & $0.264$ & $0.041$ & $0.209$ & $0.066$ & $0.050$ & $0.090$ & $0.012$ & $0.009$ \\
TabDDPM         & $0.785$ & $0.076$ & $0.473$ & $0.096$ & $0.050$ & $0.252$ & $0.574$ & $0.023$ \\
\bottomrule
\end{tabular}}
    \label{tab:privacy_full}
    \vspace{-0.6cm}
\end{table}

\vspace{-0.2cm}
\section*{Limitations and discussion}

The proposed method does not pretend to be an all-in-one solution providing high privacy and high ML utility. 
Our experiments show that TabDDPM is more private than ``shallow'' SMOTE but do not give a definite answer if TabDDPM's data can satisfy real-world privacy-concerned applications.
Therefore, the privacy problem of the DDPM-produced data needs to be further investigated.
Moreover, DCR, used in this paper, is not an ultimate privacy measure and does not cover some critical use cases. 
For example, the L2 distance between records does not consider the importance of individual features and cannot detect leakage if some sensitive features coincide.

Also, in our work, we process categorical features using multinomial diffusion. 
However, alternative approaches exist, e.g., \cite{chen2022analog, campbell2022continuous, zheng2022diffusion}.
Each of these techniques is applicable to TabDDPM and can be an interesting direction to investigate. 
As for numerical features, the possible extension of TabDDPM can be inspired by \cite{nazabal2020handling} that distinguish different types of numerical variables, i.e., real-valued, positive real-valued or ordinal.

\section{Conclusion}
\label{sect:conclusion}


In this paper, we have investigated the prospect of the diffusion modeling framework in the field of tabular data. 
In particular, we describe the DDPM design that can handle mixed data types consisting of numerical and categorical features. 
For the most considered benchmarks, the synthetic data produced by TabDDPM has consistently higher quality compared with the GAN/VAE-based rivals.
Interestingly, shallow interpolation techniques like SMOTE have demonstrated competitive ML utility and need to be considered as a simple yet effective baseline.
Nevertheless, TabDDPM outperforms SMOTE for the setups where the privacy of the data must be ensured. 

\bibliography{example_paper}

\begin{thebibliography}{52}
\providecommand{\natexlab}[1]{#1}
\providecommand{\url}[1]{\texttt{#1}}
\expandafter\ifx\csname urlstyle\endcsname\relax
  \providecommand{\doi}[1]{doi: #1}\else
  \providecommand{\doi}{doi: \begingroup \urlstyle{rm}\Url}\fi

\bibitem[Akiba et~al.(2019)Akiba, Sano, Yanase, Ohta, and
  Koyama]{akiba2019optuna}
Akiba, T., Sano, S., Yanase, T., Ohta, T., and Koyama, M.
\newblock Optuna: A next-generation hyperparameter optimization framework.
\newblock In \emph{Proceedings of the 25th ACM SIGKDD international conference
  on knowledge discovery \& data mining}, pp.\  2623--2631, 2019.

\bibitem[Austin et~al.(2021)Austin, Johnson, Ho, Tarlow, and van~den
  Berg]{austin2021structured}
Austin, J., Johnson, D.~D., Ho, J., Tarlow, D., and van~den Berg, R.
\newblock Structured denoising diffusion models in discrete state-spaces.
\newblock \emph{Advances in Neural Information Processing Systems},
  34:\penalty0 17981--17993, 2021.

\bibitem[Baldi et~al.(2014)Baldi, Sadowski, and Whiteson]{higgs}
Baldi, P., Sadowski, P., and Whiteson, D.
\newblock Searching for exotic particles in high-energy physics with deep
  learning.
\newblock \emph{Nature Communications}, 5, 2014.

\bibitem[Baranchuk et~al.(2021)Baranchuk, Rubachev, Voynov, Khrulkov, and
  Babenko]{baranchuk2021label}
Baranchuk, D., Rubachev, I., Voynov, A., Khrulkov, V., and Babenko, A.
\newblock Label-efficient semantic segmentation with diffusion models.
\newblock \emph{arXiv preprint arXiv:2112.03126}, 2021.

\bibitem[Camino et~al.(2020)Camino, Hammerschmidt,
  et~al.]{camino2020oversampling}
Camino, R.~D., Hammerschmidt, C.~A., et~al.
\newblock Oversampling tabular data with deep generative models: Is it worth
  the effort?
\newblock 2020.

\bibitem[Campbell et~al.(2022)Campbell, Benton, De~Bortoli, Rainforth,
  Deligiannidis, and Doucet]{campbell2022continuous}
Campbell, A., Benton, J., De~Bortoli, V., Rainforth, T., Deligiannidis, G., and
  Doucet, A.
\newblock A continuous time framework for discrete denoising models.
\newblock \emph{Advances in Neural Information Processing Systems},
  35:\penalty0 28266--28279, 2022.

\bibitem[Chawla et~al.(2002)Chawla, Bowyer, Hall, and
  Kegelmeyer]{chawla2002smote}
Chawla, N.~V., Bowyer, K.~W., Hall, L.~O., and Kegelmeyer, W.~P.
\newblock Smote: synthetic minority over-sampling technique.
\newblock \emph{Journal of artificial intelligence research}, 16:\penalty0
  321--357, 2002.

\bibitem[Chen et~al.(2020{\natexlab{a}})Chen, Yu, Zhang, and
  Fritz]{chen2020gan}
Chen, D., Yu, N., Zhang, Y., and Fritz, M.
\newblock Gan-leaks: A taxonomy of membership inference attacks against
  generative models.
\newblock In \emph{Proceedings of the 2020 ACM SIGSAC conference on computer
  and communications security}, pp.\  343--362, 2020{\natexlab{a}}.

\bibitem[Chen et~al.(2020{\natexlab{b}})Chen, Zhang, Zen, Weiss, Norouzi, and
  Chan]{chen2020wavegrad}
Chen, N., Zhang, Y., Zen, H., Weiss, R.~J., Norouzi, M., and Chan, W.
\newblock Wavegrad: Estimating gradients for waveform generation.
\newblock \emph{arXiv preprint arXiv:2009.00713}, 2020{\natexlab{b}}.

\bibitem[Chen et~al.(2022)Chen, Zhang, and Hinton]{chen2022analog}
Chen, T., Zhang, R., and Hinton, G.
\newblock Analog bits: Generating discrete data using diffusion models with
  self-conditioning.
\newblock \emph{arXiv preprint arXiv:2208.04202}, 2022.

\bibitem[Dhariwal \& Nichol(2021)Dhariwal and Nichol]{dhariwal2021diffusion}
Dhariwal, P. and Nichol, A.
\newblock Diffusion models beat gans on image synthesis.
\newblock 2021.

\bibitem[Engelmann \& Lessmann(2021)Engelmann and
  Lessmann]{engelmann2021conditional}
Engelmann, J. and Lessmann, S.
\newblock Conditional wasserstein gan-based oversampling of tabular data for
  imbalanced learning.
\newblock \emph{Expert Systems with Applications}, 174:\penalty0 114582, 2021.

\bibitem[Fan et~al.(2020)Fan, Liu, Li, Chen, Shen, and Du]{fan2020relational}
Fan, J., Liu, T., Li, G., Chen, J., Shen, Y., and Du, X.
\newblock Relational data synthesis using generative adversarial networks: A
  design space exploration.
\newblock \emph{arXiv preprint arXiv:2008.12763}, 2020.

\bibitem[Gorishniy et~al.(2021)Gorishniy, Rubachev, Khrulkov, and
  Babenko]{gorishniy2021revisiting}
Gorishniy, Y., Rubachev, I., Khrulkov, V., and Babenko, A.
\newblock Revisiting deep learning models for tabular data.
\newblock \emph{Advances in Neural Information Processing Systems},
  34:\penalty0 18932--18943, 2021.

\bibitem[Ho et~al.(2020)Ho, Jain, and Abbeel]{ho2020denoising}
Ho, J., Jain, A., and Abbeel, P.
\newblock Denoising diffusion probabilistic models.
\newblock 2020.

\bibitem[Hoogeboom et~al.(2021)Hoogeboom, Nielsen, Jaini, Forr{\'e}, and
  Welling]{hoogeboom2021argmax}
Hoogeboom, E., Nielsen, D., Jaini, P., Forr{\'e}, P., and Welling, M.
\newblock Argmax flows and multinomial diffusion: Learning categorical
  distributions.
\newblock \emph{Advances in Neural Information Processing Systems},
  34:\penalty0 12454--12465, 2021.

\bibitem[Hoogeboom et~al.(2022)Hoogeboom, Satorras, Vignac, and
  Welling]{hoogeboom2022equivariant}
Hoogeboom, E., Satorras, V.~G., Vignac, C., and Welling, M.
\newblock Equivariant diffusion for molecule generation in 3d.
\newblock In \emph{International Conference on Machine Learning}, pp.\
  8867--8887. PMLR, 2022.

\bibitem[Jing et~al.(2022)Jing, Corso, Chang, Barzilay, and
  Jaakkola]{jing2022torsional}
Jing, B., Corso, G., Chang, J., Barzilay, R., and Jaakkola, T.
\newblock Torsional diffusion for molecular conformer generation.
\newblock \emph{arXiv preprint arXiv:2206.01729}, 2022.

\bibitem[Jordon et~al.(2018)Jordon, Yoon, and Van Der~Schaar]{jordon2018pate}
Jordon, J., Yoon, J., and Van Der~Schaar, M.
\newblock Pate-gan: Generating synthetic data with differential privacy
  guarantees.
\newblock In \emph{International conference on learning representations}, 2018.

\bibitem[{Kelley Pace} \& Barry(1997){Kelley Pace} and Barry]{california}
{Kelley Pace}, R. and Barry, R.
\newblock Sparse spatial autoregressions.
\newblock \emph{Statistics \& Probability Letters}, 33\penalty0 (3):\penalty0
  291--297, 1997.

\bibitem[Kim et~al.(2021)Kim, Jeon, Lee, Hyeong, and Park]{kim2021oct}
Kim, J., Jeon, J., Lee, J., Hyeong, J., and Park, N.
\newblock Oct-gan: Neural ode-based conditional tabular gans.
\newblock In \emph{Proceedings of the Web Conference 2021}, pp.\  1506--1515,
  2021.

\bibitem[Kim et~al.(2022)Kim, Lee, Shin, Park, Kim, Park, and Cho]{kim2022sos}
Kim, J., Lee, C., Shin, Y., Park, S., Kim, M., Park, N., and Cho, J.
\newblock Sos: Score-based oversampling for tabular data.
\newblock In \emph{Proceedings of the 28th ACM SIGKDD Conference on Knowledge
  Discovery and Data Mining}, pp.\  762--772, 2022.

\bibitem[Kohavi(1996)]{adult}
Kohavi, R.
\newblock Scaling up the accuracy of naive-bayes classifiers: a decision-tree
  hybrid.
\newblock In \emph{KDD}, 1996.

\bibitem[Kong et~al.(2020)Kong, Ping, Huang, Zhao, and
  Catanzaro]{kong2020diffwave}
Kong, Z., Ping, W., Huang, J., Zhao, K., and Catanzaro, B.
\newblock Diffwave: A versatile diffusion model for audio synthesis.
\newblock \emph{arXiv preprint arXiv:2009.09761}, 2020.

\bibitem[Lee et~al.(2021)Lee, Hyeong, Jeon, Park, and Cho]{lee2021invertible}
Lee, J., Hyeong, J., Jeon, J., Park, N., and Cho, J.
\newblock Invertible tabular gans: Killing two birds with one stone for tabular
  data synthesis.
\newblock \emph{Advances in Neural Information Processing Systems},
  34:\penalty0 4263--4273, 2021.

\bibitem[Li et~al.(2021)Li, Yang, Chang, Feng, Xu, Li, and Chen]{li2021srdiff}
Li, H., Yang, Y., Chang, M., Feng, H., Xu, Z., Li, Q., and Chen, Y.
\newblock Srdiff: Single image super-resolution with diffusion probabilistic
  models.
\newblock 2021.

\bibitem[Li et~al.(2022)Li, Thickstun, Gulrajani, Liang, and
  Hashimoto]{li2022diffusion}
Li, X.~L., Thickstun, J., Gulrajani, I., Liang, P., and Hashimoto, T.~B.
\newblock Diffusion-lm improves controllable text generation.
\newblock \emph{arXiv preprint arXiv:2205.14217}, 2022.

\bibitem[Madeo et~al.(2013)Madeo, Lima, and Peres]{gesture}
Madeo, R. C.~B., Lima, C. A.~M., and Peres, S.~M.
\newblock Gesture unit segmentation using support vector machines: segmenting
  gestures from rest positions.
\newblock In \emph{Proceedings of the 28th Annual {ACM} Symposium on Applied
  Computing, {SAC}}, 2013.

\bibitem[Meng et~al.(2021)Meng, Song, Song, Wu, Zhu, and Ermon]{meng2021sdedit}
Meng, C., Song, Y., Song, J., Wu, J., Zhu, J.-Y., and Ermon, S.
\newblock Sdedit: Image synthesis and editing with stochastic differential
  equations.
\newblock 2021.

\bibitem[Naeem et~al.(2020)Naeem, Oh, Uh, Choi, and Yoo]{naeem2020reliable}
Naeem, M.~F., Oh, S.~J., Uh, Y., Choi, Y., and Yoo, J.
\newblock Reliable fidelity and diversity metrics for generative models.
\newblock In \emph{International Conference on Machine Learning}, pp.\
  7176--7185. PMLR, 2020.

\bibitem[Nazabal et~al.(2020)Nazabal, Olmos, Ghahramani, and
  Valera]{nazabal2020handling}
Nazabal, A., Olmos, P.~M., Ghahramani, Z., and Valera, I.
\newblock Handling incomplete heterogeneous data using vaes.
\newblock \emph{Pattern Recognition}, 107:\penalty0 107501, 2020.

\bibitem[Nichol(2021)]{nichol2021improved}
Nichol, Alex \&~Dhariwal, P.
\newblock Improved denoising diffusion probabilistic models.
\newblock \emph{ICML}, 2021.

\bibitem[Nock \& Guillame-Bert(2022)Nock and Guillame-Bert]{nock2022generative}
Nock, R. and Guillame-Bert, M.
\newblock Generative trees: Adversarial and copycat.
\newblock \emph{ICML}, 2022.

\bibitem[Pedregosa et~al.(2011)Pedregosa, Varoquaux, Gramfort, Michel, Thirion,
  Grisel, Blondel, Prettenhofer, Weiss, Dubourg, Vanderplas, Passos,
  Cournapeau, Brucher, Perrot, and Duchesnay]{scikit-learn}
Pedregosa, F., Varoquaux, G., Gramfort, A., Michel, V., Thirion, B., Grisel,
  O., Blondel, M., Prettenhofer, P., Weiss, R., Dubourg, V., Vanderplas, J.,
  Passos, A., Cournapeau, D., Brucher, M., Perrot, M., and Duchesnay, E.
\newblock Scikit-learn: Machine learning in {P}ython.
\newblock \emph{Journal of Machine Learning Research}, 12:\penalty0 2825--2830,
  2011.

\bibitem[Prokhorenkova et~al.(2018)Prokhorenkova, Gusev, Vorobev, Dorogush, and
  Gulin]{prokhorenkova2018catboost}
Prokhorenkova, L., Gusev, G., Vorobev, A., Dorogush, A.~V., and Gulin, A.
\newblock Catboost: unbiased boosting with categorical features.
\newblock \emph{Advances in neural information processing systems}, 31, 2018.

\bibitem[Rombach et~al.(2022)Rombach, Blattmann, Lorenz, Esser, and
  Ommer]{rombach2022high}
Rombach, R., Blattmann, A., Lorenz, D., Esser, P., and Ommer, B.
\newblock High-resolution image synthesis with latent diffusion models.
\newblock In \emph{Proceedings of the IEEE/CVF Conference on Computer Vision
  and Pattern Recognition}, pp.\  10684--10695, 2022.

\bibitem[Saharia et~al.(2021)Saharia, Ho, Chan, Salimans, Fleet, and
  Norouzi]{saharia2021image}
Saharia, C., Ho, J., Chan, W., Salimans, T., Fleet, D.~J., and Norouzi, M.
\newblock Image super-resolution via iterative refinement.
\newblock 2021.

\bibitem[Saharia et~al.(2022)Saharia, Chan, Saxena, Li, Whang, Denton,
  Ghasemipour, Ayan, Mahdavi, Lopes, et~al.]{saharia2022photorealistic}
Saharia, C., Chan, W., Saxena, S., Li, L., Whang, J., Denton, E., Ghasemipour,
  S. K.~S., Ayan, B.~K., Mahdavi, S.~S., Lopes, R.~G., et~al.
\newblock Photorealistic text-to-image diffusion models with deep language
  understanding.
\newblock \emph{arXiv preprint arXiv:2205.11487}, 2022.

\bibitem[Singh et~al.(2015)Singh, Sandhu, and Kumar]{fb-comments}
Singh, K., Sandhu, R.~K., and Kumar, D.
\newblock Comment volume prediction using neural networks and decision trees.
\newblock In \emph{IEEE UKSim-AMSS 17th International Conference on Computer
  Modelling and Simulation, UKSim}, 2015.

\bibitem[Sohl-Dickstein et~al.(2015)Sohl-Dickstein, Weiss, Maheswaranathan, and
  Ganguli]{sohl2015deep}
Sohl-Dickstein, J., Weiss, E., Maheswaranathan, N., and Ganguli, S.
\newblock Deep unsupervised learning using nonequilibrium thermodynamics.
\newblock In \emph{ICML}, 2015.

\bibitem[Song \& Ermon(2019)Song and Ermon]{song2019generative}
Song, Y. and Ermon, S.
\newblock Generative modeling by estimating gradients of the data distribution.
\newblock In \emph{NeurIPS}, 2019.

\bibitem[Song \& Ermon(2020)Song and Ermon]{song2020improved}
Song, Y. and Ermon, S.
\newblock Improved techniques for training score-based generative models.
\newblock \emph{NeurIPS}, 2020.

\bibitem[Song et~al.(2021)Song, Sohl-Dickstein, Kingma, Kumar, Ermon, and
  Poole]{song2020score}
Song, Y., Sohl-Dickstein, J., Kingma, D.~P., Kumar, A., Ermon, S., and Poole,
  B.
\newblock Score-based generative modeling through stochastic differential
  equations.
\newblock 2021.

\bibitem[Tashiro et~al.(2021)Tashiro, Song, Song, and Ermon]{tashiro2021csdi}
Tashiro, Y., Song, J., Song, Y., and Ermon, S.
\newblock Csdi: Conditional score-based diffusion models for probabilistic time
  series imputation.
\newblock \emph{Advances in Neural Information Processing Systems},
  34:\penalty0 24804--24816, 2021.

\bibitem[Torfi et~al.(2022)Torfi, Fox, and Reddy]{torfi2022differentially}
Torfi, A., Fox, E.~A., and Reddy, C.~K.
\newblock Differentially private synthetic medical data generation using
  convolutional gans.
\newblock \emph{Information Sciences}, 586:\penalty0 485--500, 2022.

\bibitem[Vanschoren et~al.(2014)Vanschoren, van Rijn, Bischl, and
  Torgo]{openml}
Vanschoren, J., van Rijn, J.~N., Bischl, B., and Torgo, L.
\newblock Openml: networked science in machine learning.
\newblock \emph{arXiv}, 1407.7722v1, 2014.

\bibitem[Wen et~al.(2022)Wen, Cao, Yang, Subbalakshmi, and
  Chandramouli]{wen2022causal}
Wen, B., Cao, Y., Yang, F., Subbalakshmi, K., and Chandramouli, R.
\newblock Causal-tgan: Modeling tabular data using causally-aware gan.
\newblock In \emph{ICLR Workshop on Deep Generative Models for Highly
  Structured Data}, 2022.

\bibitem[Xu et~al.(2019)Xu, Skoularidou, Cuesta-Infante, and
  Veeramachaneni]{xu2019modeling}
Xu, L., Skoularidou, M., Cuesta-Infante, A., and Veeramachaneni, K.
\newblock Modeling tabular data using conditional gan.
\newblock \emph{Advances in Neural Information Processing Systems}, 32, 2019.

\bibitem[Zhang et~al.(2021)Zhang, Zaidi, Zhou, and Li]{zhang2021ganblr}
Zhang, Y., Zaidi, N.~A., Zhou, J., and Li, G.
\newblock Ganblr: a tabular data generation model.
\newblock In \emph{2021 IEEE International Conference on Data Mining (ICDM)},
  pp.\  181--190. IEEE, 2021.

\bibitem[Zhao et~al.(2021)Zhao, Kunar, Birke, and Chen]{zhao2021ctab}
Zhao, Z., Kunar, A., Birke, R., and Chen, L.~Y.
\newblock Ctab-gan: Effective table data synthesizing.
\newblock In \emph{Asian Conference on Machine Learning}, pp.\  97--112. PMLR,
  2021.

\bibitem[Zhao et~al.(2022)Zhao, Kunar, Birke, and Chen]{zhao2022ctab}
Zhao, Z., Kunar, A., Birke, R., and Chen, L.~Y.
\newblock Ctab-gan+: Enhancing tabular data synthesis.
\newblock \emph{arXiv preprint arXiv:2204.00401}, 2022.

\bibitem[Zheng \& Charoenphakdee(2022)Zheng and
  Charoenphakdee]{zheng2022diffusion}
Zheng, S. and Charoenphakdee, N.
\newblock Diffusion models for missing value imputation in tabular data.
\newblock \emph{arXiv preprint arXiv:2210.17128}, 2022.

\end{thebibliography}
\bibliographystyle{icml2023}

\clearpage
\newpage
\appendix
\onecolumn
{\LARGE \textbf{Appendix}}
\section{MLP evaluation and tuning}
\label{app:mlp-eval-tune}
Here, we show that tuning the hyperparameters using the CatBoost guidance results in the TabDDPM models that produce synthetics that is also optimal for other classifiers/regressors. The results for a subset of datasets are presented on \autoref{tab:ml_utility_mlp}. The methods denoted with "-CB" and "-MLP" denote the CatBoost guidance and different types of evaluation (CatBoost and MLP, respectively). The "-MLP-tune" suffix stands for the MLP guidance tuning and MLP evaluation.

\begin{table}[H]
    \centering
    \caption{ML utility score with MLP evaluation and MLP tuning compared with CatBoost evaluation and CatBoost tuning. The table shows that tuning with CatBoost model provides useful synthetic for MLP.}
    {\scriptsize \begin{tabular}{lcccccccc}
\toprule
{} & AB {\tiny $\left(R2\right)$}  & AD {\tiny $\left(F1\right)$} & BU {\tiny $\left(F1\right)$} & CA {\tiny $\left(R2\right)$} & CAR {\tiny $\left(F1\right)$} & CH {\tiny $\left(F1\right)$} & DE {\tiny $\left(F1\right)$} & DI {\tiny $\left(F1\right)$} \\
\midrule
TabDDPM-CB         & ${0.550 \scriptscriptstyle \pm \scriptstyle .010}$ & ${0.795 \scriptscriptstyle \pm \scriptstyle .001}$ & ${0.906 \scriptscriptstyle \pm \scriptstyle .003}$ & $0.836 \scriptscriptstyle \pm \scriptstyle .002$ & ${0.737 \scriptscriptstyle \pm \scriptstyle .001}$ & ${0.755 \scriptscriptstyle \pm \scriptstyle .006}$ & $0.691 \scriptscriptstyle \pm \scriptstyle .004$ & ${0.740 \scriptscriptstyle \pm \scriptstyle .020}$ \\
Real-CB            & $0.556 \scriptscriptstyle \pm \scriptstyle .004$ & $0.815 \scriptscriptstyle \pm \scriptstyle .002$ & $0.906 \scriptscriptstyle \pm \scriptstyle .002$ & $0.857 \scriptscriptstyle \pm \scriptstyle .001$ & $0.738 \scriptscriptstyle \pm \scriptstyle .001$ & $0.740 \scriptscriptstyle \pm \scriptstyle .009$ & $0.688 \scriptscriptstyle \pm \scriptstyle .003$ & $0.785 \scriptscriptstyle \pm \scriptstyle .013$ \\
\midrule
TabDDPM-MLP         & ${0.569 \scriptscriptstyle \pm \scriptstyle .010}$ & ${0.794 \scriptscriptstyle \pm \scriptstyle .002}$ & ${0.903 \scriptscriptstyle \pm \scriptstyle .003}$ & ${0.809 \scriptscriptstyle \pm \scriptstyle .003}$ & ${0.737 \scriptscriptstyle \pm \scriptstyle .001}$ & ${0.750 \scriptscriptstyle \pm \scriptstyle .005}$ & ${0.679 \scriptscriptstyle \pm \scriptstyle .008}$ & ${0.754 \scriptscriptstyle \pm \scriptstyle .020}$ \\
Real-MLP           & $0.581 \scriptscriptstyle \pm \scriptstyle .005$ & $0.795 \scriptscriptstyle \pm \scriptstyle .001$ & $0.905 \scriptscriptstyle \pm \scriptstyle .003$ & $0.808 \scriptscriptstyle \pm \scriptstyle .002$ & $0.739 \scriptscriptstyle \pm \scriptstyle .001$ & $0.741 \scriptscriptstyle \pm \scriptstyle .006$ & $0.688 \scriptscriptstyle \pm \scriptstyle .004$ & $0.754 \scriptscriptstyle \pm \scriptstyle .017$ \\
\midrule
TabDDPM-MLP-tune         & $0.559 \scriptscriptstyle \pm \scriptstyle .009$ & $0.792 \scriptscriptstyle \pm \scriptstyle .002$ & $0.901 \scriptscriptstyle \pm \scriptstyle .003$ & $0.803 \scriptscriptstyle \pm \scriptstyle .004$ & $0.737 \scriptscriptstyle \pm \scriptstyle .001$ & $0.749 \scriptscriptstyle \pm \scriptstyle .006$ & $0.674 \scriptscriptstyle \pm \scriptstyle .013$ & $0.741 \scriptscriptstyle \pm \scriptstyle .018$ \\
\bottomrule
\end{tabular}
\newline
\vspace*{1em}
\newline
\begin{tabular}{lcccccccc}
\toprule
{} & FB {\tiny $\left(R2\right)$} & GE {\tiny $\left(F1\right)$} & HI {\tiny $\left(F1\right)$} & HO {\tiny $\left(R2\right)$} & IN {\tiny $\left(R2\right)$} & KI {\tiny $\left(R2\right)$} & MI {\tiny $\left(F1\right)$} & WI {\tiny $\left(F1\right)$} \\
\midrule
TabDDPM-CB         & $0.713 \scriptscriptstyle \pm \scriptstyle .002$ & $0.597 \scriptscriptstyle \pm \scriptstyle .006$ & $0.722 \scriptscriptstyle \pm \scriptstyle .001$ & ${0.677 \scriptscriptstyle \pm \scriptstyle .010}$ & $0.809 \scriptscriptstyle \pm \scriptstyle .002$ & $0.833 \scriptscriptstyle \pm \scriptstyle .014$ & ${0.936 \scriptscriptstyle \pm \scriptstyle .001}$ & $0.904 \scriptscriptstyle \pm \scriptstyle .009$ \\
Real-CB            & $0.837 \scriptscriptstyle \pm \scriptstyle .001$ & $0.636 \scriptscriptstyle \pm \scriptstyle .007$ & $0.724 \scriptscriptstyle \pm \scriptstyle .001$ & $0.662 \scriptscriptstyle \pm \scriptstyle .003$ & $0.814 \scriptscriptstyle \pm \scriptstyle .001$ & $0.907 \scriptscriptstyle \pm \scriptstyle .002$ & $0.934 \scriptscriptstyle \pm \scriptstyle .000$ & $0.898 \scriptscriptstyle \pm \scriptstyle .006$ \\
\midrule
TabDDPM-MLP         & -- & ${0.595 \scriptscriptstyle \pm \scriptstyle .006}$ & ${0.717 \scriptscriptstyle \pm \scriptstyle .002}$ & ${0.643 \scriptscriptstyle \pm \scriptstyle .010}$ & ${0.794 \scriptscriptstyle \pm \scriptstyle .008}$ & ${0.804 \scriptscriptstyle \pm \scriptstyle .015}$ & ${0.938 \scriptscriptstyle \pm \scriptstyle .001}$ & ${0.921 \scriptscriptstyle \pm \scriptstyle .006}$ \\
Real-MLP            & -- & $0.607 \scriptscriptstyle \pm \scriptstyle .007$ & $0.717 \scriptscriptstyle \pm \scriptstyle .002$ & $0.614 \scriptscriptstyle \pm \scriptstyle .006$ & $0.800 \scriptscriptstyle \pm \scriptstyle .003$ & $0.882 \scriptscriptstyle \pm \scriptstyle .004$ & $0.936 \scriptscriptstyle \pm \scriptstyle .001$& $0.905 \scriptscriptstyle \pm \scriptstyle .006$ \\
\midrule
TabDDPM-MLP-tune         & -- & -- & -- & $0.626 \scriptscriptstyle \pm \scriptstyle .009$ & $0.800 \scriptscriptstyle \pm \scriptstyle .004$ & $0.799 \scriptscriptstyle \pm \scriptstyle .018$ & -- & $0.914 \scriptscriptstyle \pm \scriptstyle .006$ \\
\bottomrule
\end{tabular}}
    \label{tab:ml_utility_mlp}
\end{table}

\section{Additional results}
\label{app:additional}

Here, we follow \cite{zhao2021ctab} and provide an additional quantitative comparison that shows how well individual feature distributions are modeled (\autoref{tab:wd_num}, \autoref{tab:js_cat}, \autoref{tab:corr_dist}). Also, we include density and coverage metrics from \cite{naeem2020reliable} that are improved alternatives of precision and recall, respectively (\autoref{tab:density}, \autoref{tab:coverage}).

\begin{table}[h!]
    \centering
    \caption{Wasserstein distance between numerical features.}
    {\scriptsize \begin{tabular}{lcccccccc}
\toprule
{} & AB & AD & BU & CA & CAR & CH & DE & DI \\
\midrule
CTGAN           & $0.008$ & $0.010$ & $0.015$ & $0.004$ & $0.004$ & $0.009$ & $0.004$ & $0.085$ \\
TVAE            & $0.020$ & $0.016$ & $0.039$ & $0.007$ & $0.027$ & $0.049$ & $0.009$ & $0.044$ \\
CTABGAN+        & $0.008$ & $0.011$ & $0.016$ & $0.019$ & $0.003$ & $0.046$ & $0.022$ & $0.016$ \\
SMOTE           & $\mathbf{0.002}$ & $0.003$ & $0.005$ & $\mathbf{0.002}$ & $0.001$ & $0.006$ & $\mathbf{0.002}$ & $0.020$ \\
TabDDPM         & $0.005$ & $\mathbf{0.002}$ & $\mathbf{0.003}$ & $\mathbf{0.002}$ & $\mathbf{0.000}$ & $\mathbf{0.005}$ & $0.012$ & $\mathbf{0.008}$ \\
\bottomrule
\end{tabular}
\begin{tabular}{lcccccccc}
\toprule
{} & FB & GE & HI & HO & IN & KI & MI & WI \\
\midrule
CTGAN           & $0.004$ & $0.010$ & $\mathbf{0.003}$ & $0.005$ & $0.021$ & $0.022$ & $0.004$ & $0.013$ \\
TVAE            & $0.008$ & $0.009$ & $0.076$ & $0.007$ & $0.025$ & $0.012$ & $0.004$ & $0.016$ \\
CTABGAN+        & $0.078$ & $0.007$ & $0.052$ & $0.008$ & $0.025$ & $0.021$ & $0.006$ & $0.006$ \\
SMOTE           & $\mathbf{0.000}$ & $\mathbf{0.004}$ & $0.009$ & $0.005$ & $0.011$ & $\mathbf{0.004}$ & $\mathbf{0.000}$ & $\mathbf{0.002}$ \\
TabDDPM         & $0.089$ & $0.011$ & $0.003$ & $\mathbf{0.004}$ & $\mathbf{0.006}$ & $0.014$ & $0.001$ & $\mathbf{0.002}$ \\
\bottomrule
\end{tabular}}
    \label{tab:wd_num}
\end{table}

\begin{table}[h!]
    \centering
    \caption{Jensen-Shannon divergence between categorical features.}
    {\scriptsize \begin{tabular}{lcccccccc}
\toprule
{} & AB & AD & BU & CA & CA & CH & DE & DI \\
\midrule
CTGAN           & $0.276$ & $0.085$ & $0.168$ & $nan$ & $0.076$ & $0.039$ & $0.120$ & $nan$ \\
TVAE            & $0.027$ & $0.095$ & $0.072$ & $nan$ & $0.181$ & $0.019$ & $0.157$ & $nan$ \\
CTABGAN+        & $0.035$ & $0.052$ & $0.037$ & $nan$ & $\mathbf{0.009}$ & $0.018$ & $0.030$ & $nan$ \\
SMOTE           & $\mathbf{0.005}$ & $0.074$ & $0.072$ & $nan$ & $0.069$ & $0.030$ & $0.058$ & $nan$ \\
TabDDPM         & $0.007$ & $\mathbf{0.019}$ & $\mathbf{0.026}$ & $nan$ & $0.011$ & $\mathbf{0.017}$ & $\mathbf{0.009}$ & $nan$ \\
\bottomrule
\end{tabular}
\begin{tabular}{lcccccccc}
\toprule
{} & FB & GE & HI & HO & IN & KI & MI & WI \\
\midrule
CTGAN           & $\mathbf{0.017}$ & $nan$ & $nan$ & $nan$ & $0.071$ & $0.296$ & $nan$ & $nan$ \\
TVAE            & $0.246$ & $nan$ & $nan$ & $nan$ & $0.033$ & $0.098$ & $nan$ & $nan$ \\
CTABGAN+        & $0.051$ & $nan$ & $nan$ & $nan$ & $0.023$ & $\mathbf{0.044}$ & $nan$ & $nan$ \\
SMOTE           & $0.027$ & $nan$ & $nan$ & $nan$ & $0.013$ & $0.102$ & $nan$ & $nan$ \\
TabDDPM         & $0.046$ & $nan$ & $nan$ & $nan$ & $\mathbf{0.008}$ & $0.060$ & $nan$ & $nan$ \\
\bottomrule
\end{tabular}}
    \label{tab:js_cat}
\end{table}

\begin{table}[h!]
    \centering
    \caption{L2 distance between correlation matrices.}
    {\scriptsize \begin{tabular}{lcccccccc}
\toprule
{} & AB & AD & BU & CA & CA & CH & DE & DI \\
\midrule
CTGAN           & $0.471$ & $0.390$ & $0.492$ & $0.606$ & $0.712$ & $0.239$ & $1.355$ & $1.735$ \\
TVAE            & $0.517$ & $0.636$ & $0.569$ & $0.753$ & $2.437$ & $0.564$ & $1.965$ & $0.736$ \\
CTABGAN+        & $0.283$ & $0.576$ & $0.164$ & $0.749$ & $0.738$ & $0.727$ & $1.496$ & $\mathbf{0.435}$ \\
SMOTE           & $\mathbf{0.185}$ & $0.482$ & $0.245$ & $0.127$ & $0.599$ & $\mathbf{0.147}$ & $\mathbf{0.642}$ & $0.838$ \\
TabDDPM         & $0.333$ & $\mathbf{0.133}$ & $\mathbf{0.068}$ & $\mathbf{0.090}$ & $\mathbf{0.202}$ & $0.161$ & $0.934$ & $0.186$ \\
\bottomrule
\end{tabular}
\begin{tabular}{lcccccccc}
\toprule
{} & FB & GE & HI & HO & IN & KI & MI & WI \\
\midrule
CTGAN           & $5.651$ & $5.301$ & $1.413$ & $0.742$ & $0.196$ & $1.530$ & $43.815$ & $0.538$ \\
TVAE            & $5.960$ & $2.996$ & $2.759$ & $0.902$ & $0.224$ & $1.004$ & $44.692$ & $0.550$ \\
CTABGAN+        & $6.782$ & $1.977$ & $1.241$ & $0.978$ & $0.207$ & $3.898$ & $31.704$ & $0.319$ \\
SMOTE           & $\mathbf{1.596}$ & $\mathbf{0.560}$ & $0.354$ & $0.452$ & $0.301$ & $\mathbf{0.569}$ & $\mathbf{0.258}$ & $\mathbf{0.059}$ \\
TabDDPM         & $16.120$ & $1.192$ & $\mathbf{0.233}$ & $\mathbf{0.336}$ & $\mathbf{0.077}$ & $3.623$ & $9.185$ & $0.375$ \\
\bottomrule
\end{tabular}}
    \label{tab:corr_dist}
\end{table}

\begin{table}[h!]
    \centering
    \caption{Density of synthetic data.}
    {\scriptsize \begin{tabular}{lcccccccc}
\toprule
{} & AB & AD & BU & CA & CA & CH & DE & DI \\
\midrule
CTGAN           & $0.224$ & $0.708$ & $0.780$ & $0.586$ & $0.938$ & $0.865$ & $0.698$ & $0.238$ \\
TVAE            & $0.347$ & $1.126$ & $1.032$ & $0.746$ & $0.845$ & $1.043$ & $0.808$ & $\mathbf{1.565}$ \\
CTABGAN+        & $0.380$ & $0.867$ & $0.998$ & $0.569$ & $0.957$ & $0.974$ & $0.730$ & $0.974$ \\
SMOTE           & $\mathbf{1.389}$ & $\mathbf{1.415}$ & $\mathbf{1.226}$ & $\mathbf{1.329}$ & $\mathbf{1.200}$ & $\mathbf{1.238}$ & $\mathbf{1.282}$ & $1.413$ \\
TabDDPM         & $0.904$ & $1.008$ & $1.116$ & $1.027$ & $1.011$ & $1.148$ & $0.810$ & $0.831$ \\
\bottomrule
\end{tabular}
\begin{tabular}{lcccccccc}
\toprule
{} & FB & GE & HI & HO & IN & KI & MI & WI \\
\midrule
CTGAN           & $0.147$ & $0.035$ & $0.702$ & $0.467$ & $0.927$ & $0.719$ & $0.361$ & $0.763$ \\
TVAE            & $0.005$ & $0.248$ & $0.960$ & $0.604$ & $1.072$ & $0.868$ & $0.747$ & $0.919$ \\
CTABGAN+        & $0.187$ & $0.448$ & $0.730$ & $0.565$ & $1.052$ & $0.186$ & $0.110$ & $0.831$ \\
SMOTE           & $\mathbf{0.926}$ & $\mathbf{1.531}$ & $\mathbf{1.682}$ & $\mathbf{1.595}$ & $\mathbf{1.213}$ & $\mathbf{1.335}$ & $\mathbf{1.308}$ & $\mathbf{1.251}$ \\
TabDDPM         & $0.633$ & $1.460$ & $1.152$ & $1.195$ & $1.150$ & $0.884$ & $0.972$ & $1.009$ \\
\bottomrule
\end{tabular}}
    \label{tab:density}
\end{table}

\begin{table}[h!]
    \centering
    \caption{Coverage of synthetic data.}
    {\scriptsize \begin{tabular}{lcccccccc}
\toprule
{} & AB & AD & BU & CA & CA & CH & DE & DI \\
\midrule
CTGAN           & $0.654$ & $0.948$ & $0.966$ & $0.759$ & $0.920$ & $1.000$ & $0.777$ & $0.572$ \\
TVAE            & $0.769$ & $0.886$ & $0.585$ & $0.922$ & $0.208$ & $0.991$ & $0.672$ & $0.978$ \\
CTABGAN+        & $0.960$ & $0.951$ & $0.999$ & $0.459$ & $0.960$ & $0.830$ & $0.841$ & $\mathbf{1.000}$ \\
SMOTE           & $\mathbf{1.000}$ & $0.970$ & $0.968$ & $\mathbf{1.000}$ & $0.866$ & $1.000$ & $0.962$ & $0.841$ \\
TabDDPM         & $\mathbf{1.000}$ & $\mathbf{0.994}$ & $\mathbf{1.000}$ & $0.998$ & $\mathbf{0.978}$ & $\mathbf{1.000}$ & $\mathbf{0.967}$ & $0.955$ \\
\bottomrule
\end{tabular}
\begin{tabular}{lcccccccc}
\toprule
{} & FB & GE & HI & HO & IN & KI & MI & WI \\
\midrule
CTGAN           & $0.238$ & $0.029$ & $0.871$ & $0.839$ & $0.986$ & $0.739$ & $0.576$ & $0.986$ \\
TVAE            & $0.014$ & $0.669$ & $0.255$ & $0.875$ & $0.987$ & $0.874$ & $0.823$ & $0.867$ \\
CTABGAN+        & $0.222$ & $0.640$ & $0.557$ & $0.952$ & $\mathbf{1.000}$ & $0.479$ & $0.241$ & $0.994$ \\
SMOTE           & $\mathbf{0.928}$ & $\mathbf{1.000}$ & $\mathbf{0.999}$ & $\mathbf{1.000}$ & $0.995$ & $0.945$ & $\mathbf{0.991}$ & $\mathbf{1.000}$ \\
TabDDPM         & $0.782$ & $0.997$ & $0.980$ & $\mathbf{1.000}$ & $\mathbf{1.000}$ & $\mathbf{0.969}$ & $0.956$ & $\mathbf{1.000}$ \\
\bottomrule
\end{tabular}}
    \label{tab:coverage}
\end{table}


\clearpage

\section{Additional visualizations}
\begin{figure*}[ht!]
\centering
\caption{The individual feature distributions for the real data and the data generated by TabDDPM, CTABGAN+, and TVAE. TabDDPM often models feature distributions more accurately than CTABGAN+ and TVAE.}
\includegraphics[width=0.80\linewidth]{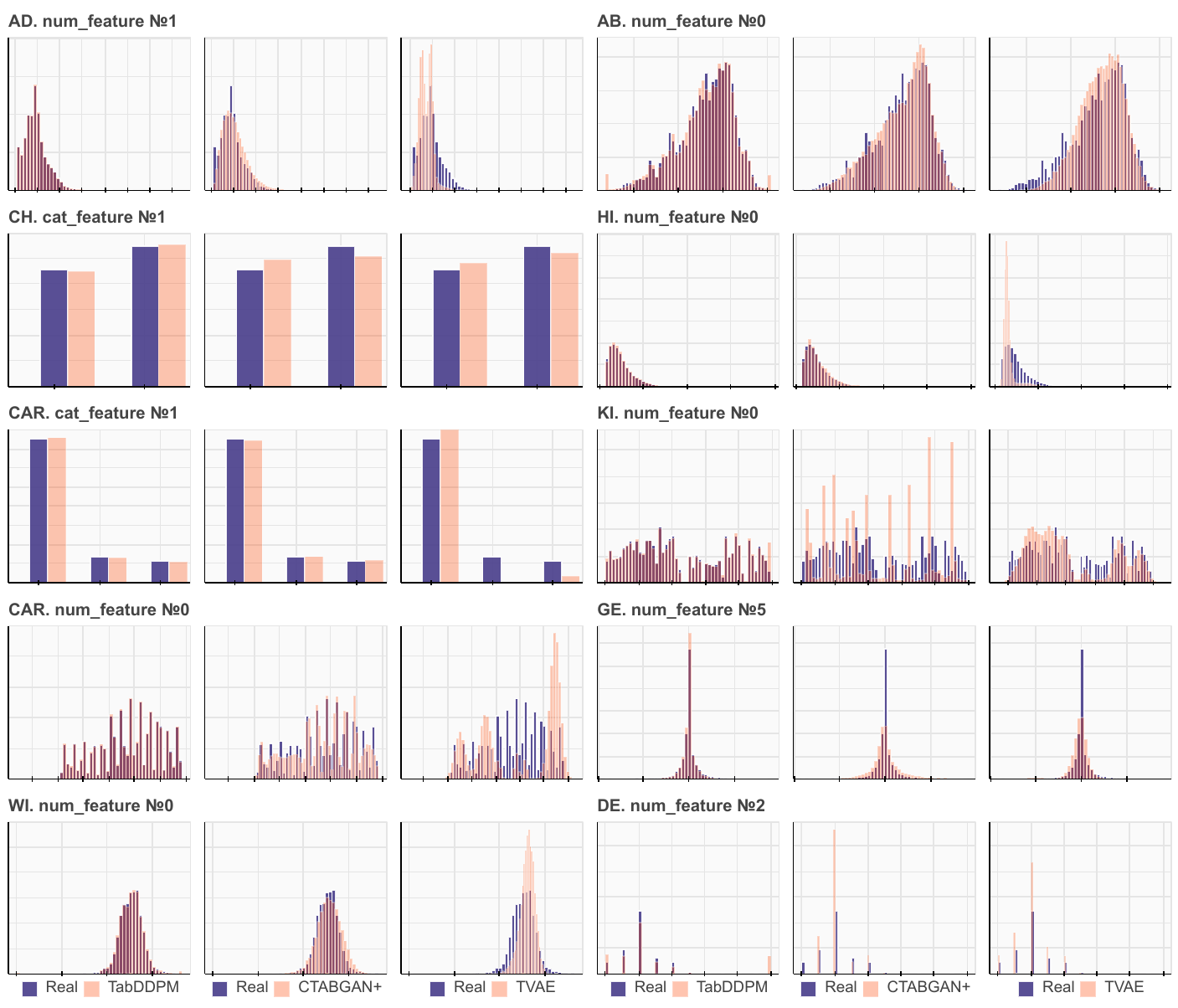}
\label{fig:features_app}
\end{figure*}

\begin{figure*}[hb!]
\centering
\caption{The absolute difference between correlation matrices computed on real and synthetic datasets. More intense red color indicates higher difference. Overall, TabDDPM captures correlations better.}
\includegraphics[width=0.80\linewidth]{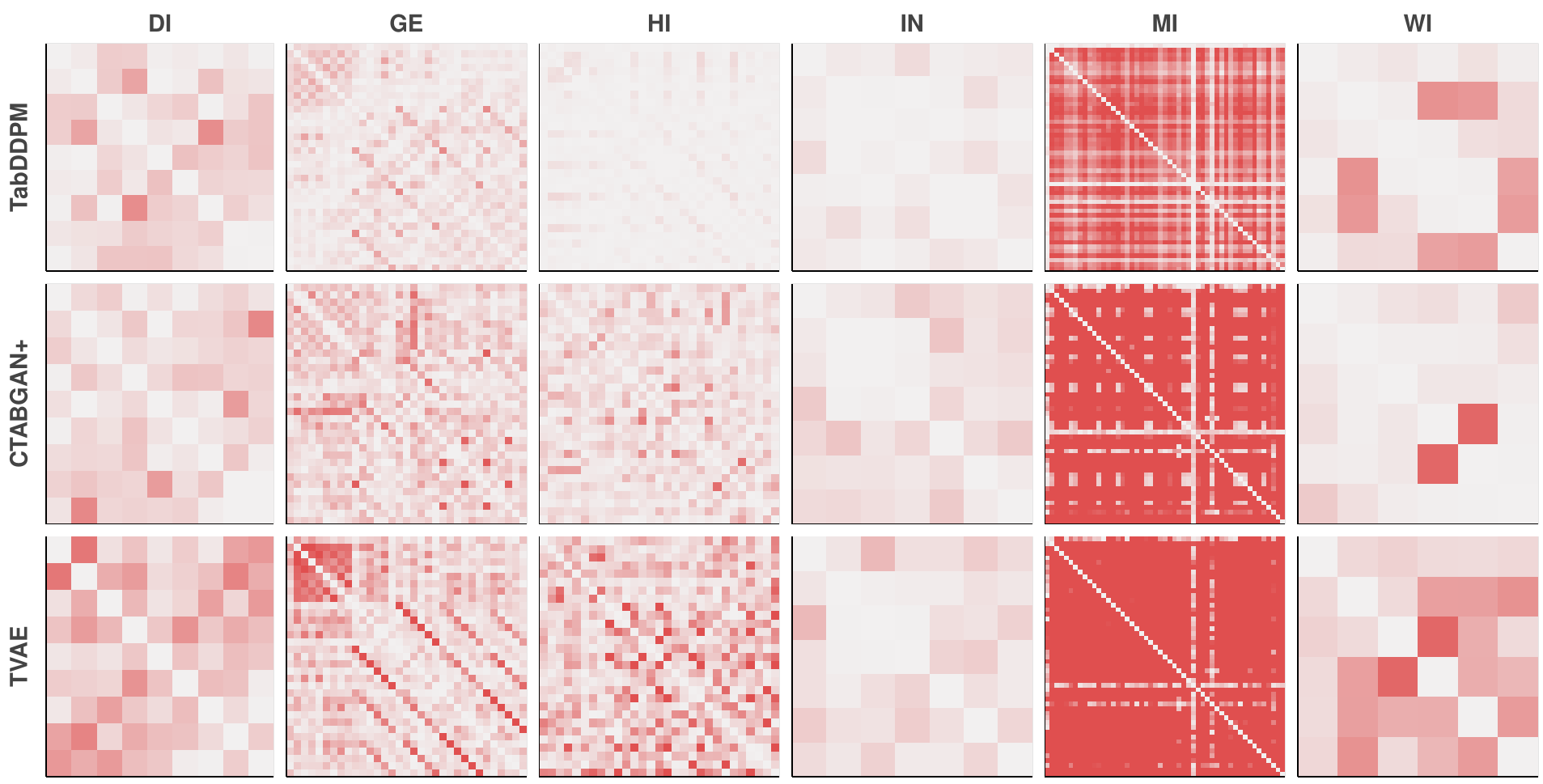}
\label{fig:heatmaps_app}
\end{figure*}

\section{Distance to Closest Record using pretrained MLP features}

This section addresses the problem that DCR in the original feature space can be an unsuitable privacy measure for SMOTE.
We pretrain the feature extractor on each dataset and compute DCR in the latent space of the MLP model.
According to the results in \autoref{tab:privacy_full_mlp}, DCR calculated on MLP features brings similar conclusions to \autoref{tab:privacy_full}.
SMOTE still significantly underperforms compared with TabDDPM.

\begin{table*}[!h]
    \centering
    \caption{Comparison in terms of mean Distance to Closest Record (DCR) calculated on pretrained MLP features (higher is better). The results are consistent with \autoref{tab:privacy_full}.}
    {\footnotesize \begin{tabular}{lcccccccc}
\toprule
{} & AB & AD & BU & CA & CAR & CH & DE & DI \\
\midrule
TVAE            & $0.282$ & $1.055$ & $0.381$ & $0.373$ & $0.173$ & $2.869$ & $0.271$ & $0.508$ \\
CTABGAN+        & $0.257$ & $1.466$ & $0.382$ & $0.332$ & $0.177$ & $2.998$ & $0.366$ & $0.669$ \\
SMOTE           & $0.081$ & $0.526$ & $0.216$ & $0.200$ & $0.147$ & $1.367$ & $0.172$ & $0.409$ \\
TabDDPM         & $0.195$ & $1.246$ & $0.330$ & $0.290$ & $0.160$ & $2.240$ & $0.168$ & $1.232$ \\
\bottomrule
\end{tabular}
\begin{tabular}{lcccccccc}
\toprule
{} & FB & GE & HI & HO & IN & KI & MI & WI \\
\midrule
TVAE            & $3.642$ & $5.484$ & $3.256$ & $0.393$ & $0.276$ & $0.513$ & $0.449$ & $0.45$ \\
CTABGAN+        & $11.44$ & $5.375$ & $4.396$ & $0.365$ & $0.305$ & $0.833$ & $12.026$ & $0.76$ \\
SMOTE           & $1.045$ & $1.673$ & $2.657$ & $0.332$ & $0.162$ & $0.294$ & $0.374$ & $0.377$ \\
TabDDPM         & $30.46$ & $3.85$ & $3.557$ & $0.336$ & $0.172$ & $0.889$ & $7.993$ & $0.620$ \\
\bottomrule
\end{tabular}}
    \label{tab:privacy_full_mlp}
\end{table*}


\section{Hyperparameters Search Spaces}
\label{sec:spaces}

\begin{table}[h!]
\centering
\small
\caption{CatBoost hyperparameters space from \cite{gorishniy2021revisiting}}
\begin{tabular}{ll}
    \toprule
    Parameter & Distribution \\
    \midrule
    Max depth & $\mathrm{UniformInt[3,10]}$ \\
    Learning rate & $\mathrm{LogUniform}[1e\text{-}5, 1]$ \\
    Bagging temperature &  $\mathrm{Uniform}[0, 1]$ \\
    L2 leaf reg  & $\mathrm{LogUniform}[1, 10]$ \\
    Leaf estimation iterations & $\mathrm{UniformInt}[1, 10]$ \\
    \midrule
    Number of tuning trials & 100 \\
    \bottomrule
\end{tabular}
\label{tab:catboost-space}
\end{table}

\begin{table}[h!]
\centering
\small
\caption{MLP hyperparameters space from \cite{gorishniy2021revisiting}}
\begin{tabular}{ll}
    \toprule
    Parameter & Distribution \\
    \midrule
    \# Layers & $\mathrm{UniformInt}[1,8]$ \\
    Layer size & $\mathrm{Int}\{64, 128, 256, 512, 1024\}$ \\
    Dropout &  $\{0, \mathrm{Uniform}[0, 0.5]\}$ \\
    Learning rate & $\mathrm{LogUniform}[1e\text{-}5, 1e\text{-}2]$ \\
    Weight decay & $\{0, \mathrm{LogUniform}[1e\text{-}6, 1e\text{-}3] \}$  \\
    \midrule
    Number of tuning trials & 100 \\
    \bottomrule
\end{tabular}
\label{tab:mlp-space}
\end{table}

\begin{table}[h!]
\centering
\small
\caption{SMOTE hyperparameters search space. $\lambda_{range}$ denotes the range of interpolation coefficient to sample from}
\begin{tabular}{ll}
    \toprule
    Parameter & Distribution \\
    \midrule
    k\_neighbours & $\mathrm{Int}[5, 20]$ \\
    $\lambda_{range}$ & $\mathrm{Float}[0, 1]$ \\
    Proportion of samples &  $\mathrm{Float}\{0.25, 0.5, 1, 2, 4, 8\}$ \\
    \midrule
    Number of tuning trials & 50 \\
    \bottomrule
\end{tabular}
\label{tab:smote-space}
\end{table}

\begin{table}[h!]
\centering
\small
\caption{CTABGAN and CTABGAN+ hyperparameters search space. See an official implementation\protect\footnotemark}
\begin{tabular}{ll}
    \toprule
    Parameter & Distribution \\
    \midrule
    \# claassif. layers & $\mathrm{UniformInt}[1,4]$ \\
    Classif. layer size & $\mathrm{Int}\{64, 128, 256\}$ \\
    Training iterations & $\mathrm{Cat}\{1000, 5000, 10000\}$ \\
    Batch Size & $\mathrm{Int}\{512, 1024, 2048\}$ \\
    random\_dim & $\mathrm{Int}\{16, 32, 64, 128\}$ \\
    num\_channels & $\mathrm{Int}\{16, 32, 64\}$ \\
    Proportion of samples &  $\mathrm{Float}\{0.25, 0.5, 1, 2, 4, 8\}$ \\
    \midrule
    Number of tuning trials & 35 \\
    \bottomrule
\end{tabular}
\label{tab:ctabgan-space}
\end{table}

\begin{table}[h!]
\centering
\small
\caption{TVAE hyperparameters search space. See an official implementation\protect\footnotemark}
\begin{tabular}{ll}
    \toprule
    Parameter & Distribution \\
    \midrule
    \# claassif. layers & $\mathrm{UniformInt}[1,6]$ \\
    Classif. layer size & $\mathrm{Int}\{64, 128, 256, 512\}$ \\
    Training iterations & $\mathrm{Cat}\{5000, 20000, 30000\}$ \\
    Batch Size & $\mathrm{Cat}\{456, 4096\}$ \\
    embedding\_dim & $\mathrm{Int}\{16, 32, 64, 128, 256, 512, 1024\}$ \\
    loss factor & $\mathrm{LogUniform}[0.01, 10]$ \\
    Proportion of samples &  $\mathrm{Float}\{0.25, 0.5, 1, 2, 4, 8\}$ \\
    \midrule
    Number of tuning trials & 50 \\
    \bottomrule
\end{tabular}
\label{tab:tvae-space}
\end{table}

\footnotetext{https://github.com/Team-TUD/CTAB-GAN-Plus}
\footnotetext{https://github.com/sdv-dev/CTGAN}

\section{Datasets}
\label{app:datasets}

We used the following datasets:
\begin{itemize}
    \item Abalone (\href{https://www.openml.org/d/183}{OpenML})
    \item Adult (income estimation, \cite{adult})
    \item Buddy (\href{https://www.kaggle.com/datasets/akash14/adopt-a-buddy}{Kaggle})
    \item California Housing (real estate data, \cite{california})
    \item Cardiovascular Disease dataset (\href{https://www.kaggle.com/datasets/sulianova/cardiovascular-disease-dataset}{Kaggle})
    \item Churn Modeling (\href{https://www.kaggle.com/shrutimechlearn/churn-modelling}{Kaggle})
    \item Diabetes (\href{https://www.openml.org/d/37}{OpenML}) 
    \item Facebook Comments Volume \cite{fb-comments} 
    \item Gesture Phase Prediction \cite{gesture}
    \item Higgs (simulated physical particles, \cite{higgs}; we use the version with 98K samples available at the OpenML repository \cite{openml})
    \item House 16H (\href{https://www.openml.org/d/574}{OpenML})
    \item Insurance (\href{https://www.kaggle.com/datasets/mirichoi0218/insurance}{Kaggle})
    \item King (\href{https://www.kaggle.com/datasets/harlfoxem/housesalesprediction}{Kaggle})
    \item MiniBooNE (\href{https://www.openml.org/d/41150}{OpenML})
    \item Wilt (\href{https://www.openml.org/d/40983}{OpenML})
\end{itemize}



\section{Environment and Runtime}
\label{app:impl}

Experiments were conducted under Ubuntu 20.04 on a machine equipped with GeForce RTX 2080 Ti GPU and Intel(R) Core(TM) i7-7800X CPU @ 3.50GHz. We used Pytorch 10.1, CUDA 11.3, scikit-learn 1.1.2 and imbalanced-learn 0.9.1 (for SMOTE).

As for runtime of the proposed method, it depends on the dataset and hyperparameters. We provide 3 examples below. All three examples use $T = 1000$ and $batch\_size=4096$. Note that hyperparameters tuning contains 50 runs and takes usually 8-10 hours. "Sample time" is for the all \textit{n\_to\_sample} number of samples.

\begin{table}[!ht]
    \centering
    \caption{Training and sampling time for TabDDPM.}
    {\scriptsize \begin{tabular}{l|c c c c|c c}
    \toprule
    Dataset & input\_dim & model\_layers & train\_steps & n\_to\_sample & train\_time & sample\_time \\ \midrule
    \texttt{CH} & 16 & [256,1024,1024, 1024,1024,512] & 30k & 26k & 670s & 6s \\  
    \texttt{HI} & 28 & [512,1024,1024, 1024,1024,512] & 30k & 502k & 502s & 430s \\
    \texttt{FB} & 146 & [512,1024] & 30k & 1264k & 783s & 470s \\ \bottomrule
\end{tabular}}
    \label{tab:runtime}
\end{table}


\end{document}